\documentclass[11pt]{article}

\usepackage[a4paper,margin=1in]{geometry}
\usepackage{times}
\usepackage[round,authoryear]{natbib}

\usepackage{amsmath,amsfonts,bm}

\def\eqref#1{equation~\ref{#1}}

\def\1{\bm{1}}

\DeclareMathAlphabet{\mathsfit}{\encodingdefault}{\sfdefault}{m}{sl}
\SetMathAlphabet{\mathsfit}{bold}{\encodingdefault}{\sfdefault}{bx}{n}

\usepackage{wrapfig}
\usepackage{booktabs}
\usepackage[table]{xcolor}
\usepackage{hyperref}
\usepackage{url}
 \usepackage{graphicx}
\usepackage{subcaption}

\usepackage{tabularx}
\usepackage{adjustbox}

\usepackage{booktabs}
\usepackage[table]{xcolor}
\usepackage{colortbl}
\usepackage{xfp}

\definecolor{heatblue}{RGB}{24,95,156}

\newcommand{\heatcell}[2]{%
  \edef\normscore{%
    \fpeval{max(0,min(1,(#1-#2)/(100-#2)))}%
  }%
  \edef\shade{%
    \fpeval{
      round(
        max(2,min(96,
          2
          + 88*(\normscore)^(1.2)
          + 24*(max(0,(\normscore-0.65)/0.35))^(1.25)
        )),0
      )
    }%
  }%
  \edef\cellcmd{\noexpand\cellcolor{heatblue!\shade!white}}%
  \cellcmd #1%
}

\newcommand{\heatbest}[2]{%
  \edef\normscore{%
    \fpeval{max(0,min(1,(#1-#2)/(100-#2)))}%
  }%
  \edef\shade{%
    \fpeval{
      round(
        max(2,min(96,
          2
          + 88*(\normscore)^(1.2)
          + 24*(max(0,(\normscore-0.65)/0.35))^(1.25)
        )),0
      )
    }%
  }%
  \edef\cellcmd{\noexpand\cellcolor{heatblue!\shade!white}}%
  \cellcmd \textbf{#1}%
}

\begin{document}


\begin{flushleft}

{\fontsize{19}{23}\selectfont\scshape
Every Model Leaves a Trace:\\
Multi-Resolution LLM Fingerprinting
\par}

\vspace{1.3em}


{\normalsize\bfseries
Mohammad Reza Modarres\textsuperscript{*1,2}
\quad
Armin Tourajmehr\textsuperscript{*2}\\
Artin Tavasoli\textsuperscript{1}
\quad
Yadollah Yaghoobzadeh\textsuperscript{$\dagger$1,2}\\
Mohammad Taher Pilehvar\textsuperscript{$\dagger$3}
\par}

\vspace{0.3em}


{\small
\textsuperscript{1}School of Electrical and Computer Engineering,\\
College of Engineering, University of Tehran, Tehran, Iran\\
\textsuperscript{2}Tehran Institute for Advanced Studies, Khatam University, Iran\\
\textsuperscript{3}Cardiff University, United Kingdom
\par}

\vspace{0.3em}


{\small\ttfamily\color{blue}
mr.modarres@ut.ac.ir,
a.tourajmehr@teias.institute\\
artin.tavassoli@ut.ac.ir,
y.yaghoobzadeh@ut.ac.ir\\
pilehvarmt@cardiff.ac.uk
\par}

\vspace{0.6em}

{\footnotesize
$^*$Equal contribution.
$\dagger$Corresponding authors.
\par}

\end{flushleft}

\vspace{1.5em}


\begin{center}
\textbf{{\large\scshape Abstract}}
\end{center}

\vspace{0.3em}

\begin{center}
\begin{minipage}{0.86\textwidth}
\normalsize

Large language models leave model-specific traces in their
generated text, but it remains unclear how finely these
fingerprints can distinguish related models and which
signals remain informative as obvious cues are reduced.
We study this through \emph{fingerprint resolution},
considering both model resolution, from different families
to distilled and adapted variants, and signal resolution
across behavioral, stylometric, lexical, and semantic
evidence. We progressively suppress surface cues,
paraphrase outputs, vary the amount of available text,
and test generalization across prompt distributions
and languages. Our results show that fingerprints persist
beyond broad family differences, remaining detectable
in distilled models and even LoRA adaptations of the
same base model. Although attribution weakens as
surface and lexical cues are altered, substantial
model-specific signal remains. These signals transfer
across prompt sources and languages, with semantic
representations proving most robust across languages.
These findings show that model identity is recoverable
at multiple resolutions, but that recoverability depends
on model similarity and the evidence available in the text.

\end{minipage}
\end{center}

\section{Introduction}
\label{sec:introduction}

LLMs often leave recognizable traces in their generated text, making it possible to distinguish among different source models~\citep{uchendu-etal-2020-authorship, rao2025birdsstonemultitaskdetection, antoun2024textsourceresultsdetecting}. Some traces are easy to observe when a model generates text, such as producing longer responses, using more headings, following particular formatting conventions, or beginning answers in recurring ways. Such signals can make attribution relatively easy, but they are also fragile, and post-generation editing can weaken text-based detectors~\citep{krishna2023paraphrasing, dugan-etal-2024-raid}.
What remains unclear is which model-specific signals survive when these visible cues are reduced, especially when the candidate models are closely related. We therefore ask: \emph{how finely can the identity of an LLM be resolved from its generated text, and which properties of the text still reveal that identity?}

This question is increasingly important as modern LLMs become more diverse and increasingly appear as closely related model families rather than isolated systems. Different sizes, distilled descendants, and adapted variants may share architectures, training histories, or development processes while differing only in subtle ways, making source attribution more difficult. At the same time, generated text may be reformatted, edited, shortened, or paraphrased before attribution, weakening or removing model-identifying signals. Attribution must therefore address two related challenges by distinguishing increasingly similar models and determining which identifying signals remain observable after the text is transformed.

\begin{figure*}[t]
    \centering
    \includegraphics[
        width=\textwidth,
        keepaspectratio
    ]{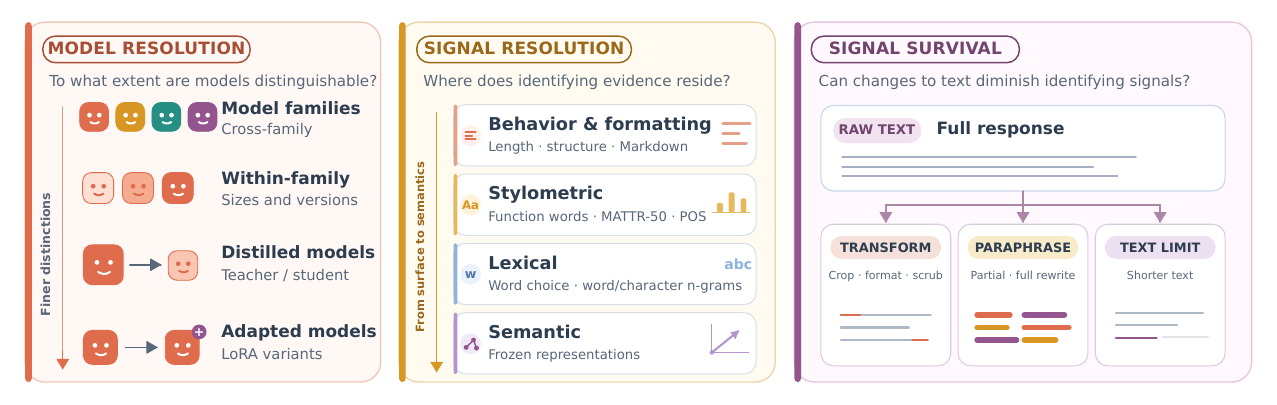}
    \caption{
Overview of the fingerprint-resolution framework, linking model similarity,
observable evidence, and generalization across prompts, languages, and text budgets.
}
    \label{fig:fingerprint_resolution}
\end{figure*}

We study this problem through \emph{fingerprint resolution}, illustrated in Figure~\ref{fig:fingerprint_resolution}. \emph{Model resolution} describes how attribution changes as candidate generators become more similar, from different families to within-family models, distilled descendants, and controlled LoRA~\citep{hu2022lora} adaptations. \emph{Signal resolution} describes which behavioral, stylometric, lexical, and semantic signals carry model identity and whether they survive as observable cues are reduced. Together, these complementary axes distinguish the existence of a fingerprint from the conditions under which it remains detectable.

Prior work has shown that source models can be identified using lexical and grammatical patterns, learned representations, and supervised attribution models~\citep{mcgovern-etal-2025-large,sun2025idiosyncrasies,lu-etal-2026-synergizing}. Our goal is to use model attribution as a diagnostic tool rather than treat accuracy as the endpoint. We progressively remove observable cues, paraphrase model outputs, and control the amount of available text to determine which behavioral, stylometric, lexical, and semantic signals continue to reveal model identity.
We further test whether fingerprints generalize beyond the conditions under which they are learned. Source-held-out experiments measure transfer across prompt distributions, while within- and cross-language experiments evaluate English, Chinese, and Persian. These settings address evidence that source attribution may weaken across languages~\citep{la-cava-etal-2026-authorship} and that fingerprint visibility can depend on language~\citep{gloaguen2026llm}.


Our results show that model fingerprints extend beyond broad family differences and remain detectable even between distilled models and their teachers, as well as between LoRA-adapted versions of the same base model.
Attribution becomes harder as behavioral and stylometric cues are removed and lexical choices are altered through paraphrasing. Cross-lingual experiments show that semantic representations provide the most reliable transfer across languages, while cross-dataset results show substantial transfer across prompt sources but sharp degradation for highly constrained generation tasks. Model identity is therefore recoverable at different resolutions depending on model similarity and the evidence available in generated text.

Our key contributions are threefold: (1) \textbf{A fingerprint-resolution framework} spanning different levels of model similarity, revealing distinguishable fingerprints even in distilled models and LoRA-adapted variants of the same base model; (2) \textbf{A decomposition of model-identifying evidence} using multiple signal families, progressive cue suppression, paraphrasing, and varying the amount of available text; and (3) \textbf{A broad evaluation of fingerprint generalization} across unseen prompt distributions, different generation tasks, and within- and cross-language settings in English, Chinese, and Persian.

\newcommand{\writenote}[1]{%
    \noindent\textit{\small Writing note: #1}\par\medskip
}

\section{Related Work}
\label{sec:related_work}
Prior work has shown that LLM-generated text contains model-specific signals that can be used to identify its source. These signals appear at different levels, from broad differences associated with model family and size~\citep{antoun2024textsourceresultsdetecting} to finer patterns in vocabulary, character use, and sentence structure that can persist across domains~\citep{mcgovern-etal-2025-large}. Other approaches use learned representations, contrastive learning, or specialized attribution models to improve source classification~\citep{sun2025idiosyncrasies, bitton2025detecting,lu-etal-2026-synergizing, guo2024detective, la-cava-tagarelli-2025-openturingbench}. Distinguishing signals can even emerge among closely related models that differ only in training details such as initialization or data order~\citep{suzuki2025natural,tong2026seedprints}. 
Together, these studies establish that model identity is often recoverable from generated text. Less clear is how that recoverability changes as candidate models become more similar and as the available identifying cues are progressively removed.

Several studies have also examined the robustness of model attribution. Some control for response length and formatting or apply transformations such as translation and summarization~\citep{sun2025idiosyncrasies,mcgovern-etal-2025-large}. Related detection benchmarks show that robustness can deteriorate when outputs come from previously unseen generators~\citep{li-etal-2024-mage, wang-etal-2024-m4}.
Ownership-oriented methods take a different approach by implanting private trigger-response behaviors or probing models with selected prompts and repeated observations~\citep{xu-etal-2024-instructional,pasquini2025llmmap,gao2025equality}. Our setting is complementary. Rather than optimizing prompts or attribution models, we study passive attribution from ordinary model outputs and ask how much identifying information remains when text is shortened, reformatted, partly rewritten, or paraphrased. This motivates our view of fingerprinting as a \emph{resolution} problem, determined jointly by the similarity of the candidate models and the amount and type of evidence that remains observable.

A smaller body of work considers multilingual attribution. Recent results suggest that source-model attribution does not always transfer reliably across languages~\citep{la-cava-etal-2026-authorship,greco2026multighostbenchmultilingualbenchmarklongform}, while semantically conditioned watermarking shows that language can be used as a channel for embedding model-specific signals~\citep{gloaguen2026llm}. In contrast, we focus on naturally occurring model fingerprints rather than deliberately embedded signals, and examine which behavioral, lexical, and semantic cues remain informative within and across English, Chinese, and Persian.

\section{Experimental Setup}
\label{sec:setup}

As discussed in Section~\ref{sec:related_work}, most prior work frames attribution as a closed-set classification task in which a generated response is assigned to one of several predefined candidate models and performance is summarized by a single score. While this formulation establishes whether model-specific signals are recoverable, it provides limited insight into how attribution changes as candidate models become increasingly similar or which properties of the generated text support the prediction. These questions matter because modern LLMs often appear as closely related variants and because their outputs may be altered before attribution. We study these factors through \emph{fingerprint resolution}.

In this section, we first formalize attribution as a measure of fingerprint recoverability and then define model resolution (Sec.~\ref{sec:models}), signal representations (Sec.~\ref{sec:signals}), and text transformations (Sec.~\ref{sec:transformations}) used to study both axes of fingerprint resolution.
We then study these two axes jointly in Sec.~\ref{sec:results} and examine whether the resulting fingerprints generalize across prompt distributions and languages in Sec.~\ref{sec:generalization}. Detailed generation, feature, and classifier settings are provided in Appendix~\ref{app:experimental-details}.


\paragraph{Problem Formulation.}
Let $\mathcal{M}=\{M_1,\ldots,M_K\}$ denote the set of candidate generators. Given a response $y$, the attribution task predicts its source model as $\hat{M}=f(y)$. Rather than treating the performance of $f$ as the endpoint, we use it as a measure of fingerprint recoverability. We examine how attribution changes along two complementary axes: \emph{model resolution}, which varies how closely related the candidate generators are, and \emph{signal resolution}, which varies the type and amount of model-identifying evidence available in their outputs.

\subsection{Model Resolution}
\label{sec:models}

Model resolution is examined across four levels of similarity between candidate generators. Our primary model set contains 19 models from eight families and supports the cross-family and within-family experiments. Additional models are used for the distillation and controlled-adaptation settings. Complete model lists and exact identifiers are provided in Appendix~\ref{app:generation}. Across settings, we keep the attribution protocol as consistent as possible so that differences in performance primarily reflect the similarity between candidate models.

\begin{itemize}
\item \textbf{Cross-Family.}
We select one representative from each of the eight families: Llama-4 Maverick, Qwen-3 32B, Mistral-Small-3.2 24B, Gemma-4 31B, DeepSeek-V4 Flash-0731, Claude Haiku 4.5, Gemini-2.5 Flash, and GPT-4.1. 

\item \textbf{Within-Family.}
We compare all model pairs within families containing multiple models. This gives six Llama pairs, three Qwen pairs, and one pair each for other model families.

\item \textbf{Distillation Relationships.}
We examine three teacher--student pairs: Llama-3.1 8B vs. Llama-3.1-Minitron 4B, Mistral-NeMo 12B vs. Mistral-NeMo-Minitron 8B, and Gemma-2 27B vs. Gemma-2 2B. 

\item \textbf{Controlled Adaptations.}
We use Llama-3.1-8B-Instruct and Qwen3-8B as base models and fine-tune each on GSM8K~\citep{cobbe2021trainingverifierssolvemath} and MATH~\citep{hendrycks2021measuringmathematicalproblemsolving} using LoRA. We then vary the strength of the learned LoRA update to create a controlled sequence of adaptations and test when their outputs become distinguishable from those of the corresponding base model. The fine-tuning procedure and LoRA settings are described in 
Appendix~\ref{app:lora_res}.

\end{itemize}

\subsection{Signal Resolution: From Surface to Semantics}
\label{sec:signals}

Signal resolution covers four complementary levels of model-identifying signals, ranging from surface-level behavioral patterns to semantic representations.

\begin{itemize}
\item \textbf{Behavioral and Structural.}
These features capture how a model organizes and presents its response. They cover response length and structure, formatting and layout, punctuation, and repetition tendency. Representative examples include sentence and paragraph counts, headings and lists, punctuation usage, and bigram and trigram repetition ratios.

\item \textbf{Stylometric.}
These features capture recurring patterns in vocabulary use and grammatical structure. They cover vocabulary diversity, word-form statistics, vocabulary distribution, function and grammatical words, repetition, and the ratio of numeric tokens. We concatenate 17 dense lexical features with POS (part-of-speech) unigram, bigram, and trigram TF--IDF representations, inspired by prior attribution work~\citep{mcgovern-etal-2025-large}.

\item \textbf{Lexical.}
To capture lexical form directly, we use TF--IDF representations at both the word and character levels. The word-level representation uses unigrams and bigrams, while the character-level representation uses 3--5-grams. We concatenate these two representations into a single lexical feature set.

\item \textbf{Semantic.}
We use \texttt{openai/text-embedding-3-small} to obtain a fixed embedding for each response. This provides a semantic-level view that can capture model-specific information not explicitly represented by the manually defined or surface-form features.

\end{itemize}

Two additional derived representations are considered. First, for the \textbf{prompt-neutralized embedding}, we compute the mean embedding across all candidate-model responses to the same prompt and subtract this mean from each response embedding. This removes variation shared across models for that prompt and emphasizes differences between their responses. Because this setting requires peer responses from all candidate models for the same test prompt, it differs from standard attribution and is used as a diagnostic representation. If attribution improves after neutralization, it suggests that removing prompt- and task-shared variation exposes model-specific signals that are more discriminative and less dependent on the prompt itself. Second, the \textbf{combined} representation concatenates the four main signal groups described above to capture their complementary information. Complete feature definitions, vectorizer settings, and dimensionality limits are provided in Appendix~\ref{app:features}.


\begin{table}[t]
\centering
\small
\setlength{\tabcolsep}{5pt}
\renewcommand{\arraystretch}{1.08}

\begin{tabular}{cl}
\toprule
\textbf{View} & \textbf{Transformation} \\
\midrule
$T_0$ & Raw response \\
$T_1$ & Remove words from the beginning and end \\
$T_2$ & $T_1$ + remove formatting and layout cues \\
$T_3$ & $T_2$ + lowercase and remove punctuation, symbols, and emojis \\
\midrule
$P_1$ & Paraphrase approximately one third of eligible sentences in $T_0$ \\
$P_2$ & Paraphrase the full $T_0$ response \\
$P_1^*$ & Apply $T_3$ transformation to $P_1$ \\
$P_2^*$ & Apply $T_3$ transformation to $P_2$ \\
\bottomrule
\end{tabular}

\caption{Response views used to progressively remove or rewrite model-identifying cues.}
\label{tab:text_views}
\end{table}


\subsection{Transformations}
\label{sec:transformations}
Attribution is evaluated as candidate models become more similar and as identifying signals are progressively weakened. Table~\ref{tab:text_views} summarizes the response views used throughout the experiments. $T_1$--$T_3$ form a cumulative signal-removal sequence, while $P_1$ and $P_2$ are generated directly from the raw response through LLM-based partial and full paraphrasing, respectively. We then apply $T_3$ to $P_1$ and $P_2$ to obtain $P_1^*$ and $P_2^*$, allowing us to test paraphrased text after the strongest surface normalization. Full transformation and paraphrasing details are provided in Appendix~\ref{app:transformation-detail}.

\begin{table}[t]
\centering
\small
\setlength{\tabcolsep}{5pt}
\renewcommand{\arraystretch}{1.15}

\begin{tabular}{lccccc@{\hspace{7pt}}c}
\toprule
\textbf{View} &
\textbf{Behavioral} &
\textbf{Stylometric} &
\textbf{Lexical} &
\textbf{Semantic} &
\textbf{Combined} &
\textbf{PN Embedding\textsuperscript{\textdagger}} \\
\midrule

T0 &
\heatcell{52.42}{12.5} &
\heatcell{60.72}{12.5} &
\heatcell{80.07}{12.5} &
\heatcell{66.04}{12.5} &
\heatbest{81.41}{12.5} &
\heatcell{83.75}{12.5} \\

T1 &
\heatcell{48.78}{12.5} &
\heatcell{58.93}{12.5} &
\heatbest{76.27}{12.5} &
\heatcell{60.81}{12.5} &
\heatcell{76.14}{12.5} &
\heatcell{76.13}{12.5} \\

T2 &
\heatcell{41.88}{12.5} &
\heatcell{56.34}{12.5} &
\heatbest{72.48}{12.5} &
\heatcell{56.23}{12.5} &
\heatcell{71.86}{12.5} &
\heatcell{74.42}{12.5} \\

T3 &
\heatcell{25.59}{12.5} &
\heatcell{49.62}{12.5} &
\heatbest{67.76}{12.5} &
\heatcell{50.94}{12.5} &
\heatcell{67.04}{12.5} &
\heatcell{69.35}{12.5} \\

\addlinespace[2pt]

P1 &
\heatcell{50.86}{12.5} &
\heatcell{58.20}{12.5} &
\heatcell{74.42}{12.5} &
\heatcell{63.69}{12.5} &
\heatbest{77.85}{12.5} &
\heatcell{80.90}{12.5} \\

P2 &
\heatcell{48.12}{12.5} &
\heatcell{53.32}{12.5} &
\heatcell{68.18}{12.5} &
\heatcell{57.81}{12.5} &
\heatbest{71.86}{12.5} &
\heatcell{75.23}{12.5} \\

\addlinespace[2pt]

P1$^\ast$ &
\heatcell{24.48}{12.5} &
\heatcell{46.51}{12.5} &
\heatcell{61.01}{12.5} &
\heatcell{47.51}{12.5} &
\heatbest{61.47}{12.5} &
\heatcell{64.99}{12.5} \\

P2$^\ast$ &
\heatcell{23.08}{12.5} &
\heatcell{41.79}{12.5} &
\heatcell{53.94}{12.5} &
\heatcell{41.90}{12.5} &
\heatbest{55.87}{12.5} &
\heatcell{59.39}{12.5} \\

\bottomrule
\end{tabular}

\caption{Cross-family attribution accuracy (\%) using logistic regression. Chance accuracy is 12.5\%.}
\textsuperscript{\textdagger}PN Embedding denotes the Prompt-Neutralized Embedding accuracy. This representation requires access to peer responses from all candidate models for the same prompt. We use ``PN Embedding'' and this symbol consistently throughout the paper.
\label{tab:cross_family}
\end{table}




\subsection{Datasets and Evaluation}
\label{sec:data}
To make the attribution setting more realistic, we use open-ended prompts and sample responses at temperature \(1.0\), producing substantial generations; approximately \(95\%\) of responses exceed 50 words. The main English experiments use four prompt sets, \(D_1\)--\(D_4\), with 200 prompts from each source before filtering. \(D_1\) and \(D_2\) are independent samples from INFINITY-CHAT~\citep{jiang2026artificial}, \(D_3\) contains researcher-designed prompts spanning five task categories (construction described in Appendix~\ref{app:d3-construction}), and \(D_4\) comes from a separate general-prompt source~\citep{no_robots}. After removing one near-duplicate, the final set contains 799 prompts. All models follow a common decoding protocol (Appendix~\ref{app:gen-conf}). The multilingual experiments in Sec.~\ref{sec:generalization} extend this setup to English, Mandarin Chinese, and Persian by translating the English prompts with GPT-4o-mini.

For attribution, we use Logistic Regression (LR) as the primary diagnostic classifier and a one-hidden-layer MLP as a nonlinear robustness check. The main English experiments use five-fold, prompt-grouped, dataset-stratified cross-validation, keeping all outputs and response views associated with the same prompt within a single fold. We reuse the same fold assignments across experiments and report accuracy. Classifier settings and evaluation details are provided in Appendix~\ref{app:classifier-details}.

\section{Resolving Model Fingerprints}
\label{sec:results}

This section presents the results of the \emph{model- and signal-resolution} experiments. Attribution is evaluated across different signal types while observable signals are progressively reduced to test whether model fingerprints persist. The analysis also spans increasing levels of model similarity, from distinct model families to closely related adaptations of the same base model. Neither LR nor the MLP consistently outperforms the other across settings. We therefore report LR results throughout the main text and provide the corresponding MLP results in the Appendices \ref{app:cross-family-mlp} and \ref{app:within_other_families}.

\paragraph{Across Model Families.}
Table~\ref{tab:cross_family} presents the eight-way cross-family attribution results, with a chance level of 12.5\%. The results show that attribution relies on more than formatting signals and response templates. Behavioral features show the largest relative decline under edge cropping (T1) and formatting removal (T2), consistent with their direct encoding of response structure, while stylometric and lexical representations remain more stable. Even after T3 removes formatting, capitalization, punctuation, symbols, and emojis, the lexical representation remains the strongest individual signal group. This suggests that model-specific word choices and local lexical patterns persist after visible surface signals are suppressed. Embedding accuracy also declines under T1--T3, indicating that frozen representations capture some of this affected lexical and stylistic information rather than semantic content alone.

Paraphrasing produces a complementary pattern. Accuracy decreases from P1 to P2 as more of the original text is rewritten, and falls further in P1$^\ast$ and P2$^\ast$ after applying T3. Combined performs best on the raw and paraphrased views, whereas the Lexical representation slightly exceeds it under T1--T3. This shift is consistent with paraphrasing weakening lexical realization while T3 removes additional structural and surface signals. Even under P2$^\ast$, all representations remain above chance, with Combined reaching 55.87\%. Substantial cross-family information therefore remains recoverable after both lexical realization and visible surface form are heavily altered.

The Prompt-Neutralized Embedding (PN Embedding) achieves the highest accuracy across all views. We report it separately because it is a diagnostic, prompt-relative representation rather than a standard standalone classifier input. Unlike the other representations, PN Embedding requires peer responses from all candidate models to the same prompt so that their shared embedding mean can be removed. Its strong performance suggests that removing prompt- and task-shared variation exposes model-specific signals that are more discriminative and less dependent on the prompt itself. This finding also motivates the analysis of generalization across prompt distributions in Sec.~\ref{sec:generalization}. Corresponding results for the MLP classifier are provided in Appendix~\ref{app:cross-family-mlp}. Across the reported settings, the largest standard deviation is 3.2\%, indicating relatively low variability. To avoid overloading the tables, we omit standard deviations from the main presentation.

\begin{table}[t]
\centering
\small
\setlength{\tabcolsep}{5pt}
\renewcommand{\arraystretch}{1.15}

\begin{adjustbox}{max width=\linewidth}
\begin{tabular}{lccccc@{\hspace{7pt}}c}
\toprule
\textbf{Model Pair} &
\textbf{Behavioral} &
\textbf{Stylometric} &
\textbf{Lexical} &
\textbf{Semantic} &
\textbf{Combined} &
\textbf{PN Embedding\textsuperscript{\textdagger}} \\
\midrule

3.1-70B vs.\ 3.3-70B
& \heatcell{60.33}{50}
& \heatcell{59.45}{50}
& \heatcell{59.76}{50}
& \heatcell{59.70}{50}
& \heatbest{66.40}{50}
& \heatcell{73.34}{50} \\

3.1-70B vs.\ 4-Maverick
& \heatcell{60.45}{50}
& \heatcell{63.33}{50}
& \heatcell{69.02}{50}
& \heatcell{60.51}{50}
& \heatbest{72.09}{50}
& \heatcell{76.35}{50} \\

3.1-8B vs.\ 3.1-70B
& \heatcell{55.38}{50}
& \heatcell{58.26}{50}
& \heatcell{59.01}{50}
& \heatcell{55.38}{50}
& \heatbest{61.39}{50}
& \heatcell{63.33}{50} \\

3.1-8B vs.\ 3.3-70B
& \heatcell{57.51}{50}
& \heatcell{61.51}{50}
& \heatcell{60.76}{50}
& \heatcell{58.64}{50}
& \heatbest{64.27}{50}
& \heatcell{68.21}{50} \\

3.1-8B vs.\ 4-Maverick
& \heatcell{56.32}{50}
& \heatcell{62.83}{50}
& \heatcell{68.59}{50}
& \heatcell{58.57}{50}
& \heatbest{70.96}{50}
& \heatcell{69.71}{50} \\

3.3-70B vs.\ 4-Maverick
& \heatcell{58.64}{50}
& \heatcell{61.64}{50}
& \heatbest{68.02}{50}
& \heatcell{57.76}{50}
& \heatcell{67.58}{50}
& \heatcell{73.22}{50} \\

\bottomrule
\end{tabular}
\end{adjustbox}

\caption{Within-family attribution accuracy (\%) for Llama model pairs using logistic regression. Each row is a binary attribution task, with chance accuracy of 50\%.}
\label{tab:within_llama}
\end{table}

\paragraph{Within Model Families.}
Table~\ref{tab:within_llama} shows that model identity remains detectable among closely related Llama models, although accuracy is substantially lower than in the cross-family setting. The Combined representation performs best for most pairs, suggesting that distinctions between related models draw on multiple complementary signal groups. The Lexical representation is generally the strongest individual signal group, while behavioral and embedding representations provide weaker separation. Pairwise difficulty is not determined by parameter count alone: differences across model generations are often easier to detect than size differences within the same generation. Overall, within-family attribution appears to rely on several weak but complementary signals rather than a single dominant fingerprint. Full transformation results for the Llama within-family pairs are provided in Appendix~\ref{app:llama_pairwise_views}.

The same broad pattern extends to the other model families, although fingerprint strength varies considerably across them. Lexical or Combined representations are typically strongest, while behavioral and embedding representations alone are less reliable. Full results are reported in Appendix~\ref{app:within_other_families}.

\begin{table}[t]
\centering
\small
\setlength{\tabcolsep}{5pt}
\renewcommand{\arraystretch}{1.15}

\begin{adjustbox}{max width=\linewidth}
\begin{tabular}{lccccc@{\hspace{7pt}}c}
\toprule
\textbf{Teacher $\rightarrow$ Student} &
\textbf{Behavioral} &
\textbf{Stylometric} &
\textbf{Lexical} &
\textbf{Semantic} &
\textbf{Combined} &
\textbf{PN Embedding\textsuperscript{\textdagger}} \\
\midrule

Llama 8B $\rightarrow$ Minitron 4B
& \heatcell{66.67}{50}
& \heatcell{69.49}{50}
& \heatcell{71.24}{50}
& \heatcell{70.68}{50}
& \heatbest{74.56}{50}
& \heatcell{80.58}{50} \\

Mistral 12B $\rightarrow$ Minitron 8B
& \heatcell{99.31}{50}
& \heatcell{99.44}{50}
& \heatcell{99.12}{50}
& \heatbest{99.62}{50}
& \heatbest{99.62}{50}
& \heatcell{99.87}{50} \\

Gemma 27B $\rightarrow$ Gemma 2B
& \heatcell{69.92}{50}
& \heatcell{67.98}{50}
& \heatcell{69.80}{50}
& \heatcell{72.74}{50}
& \heatbest{77.94}{50}
& \heatcell{88.47}{50} \\

\bottomrule
\end{tabular}
\end{adjustbox}

\caption{Teacher--student attribution accuracy (\%) using logistic regression. Each row is a binary classification task with a chance level of 50\%.}
\label{tab:distillation_attribution}
\end{table}

\paragraph{Teacher--Student Attribution.}
Table~\ref{tab:distillation_attribution} shows that distilled students remain readily distinguishable from their teachers. Compared with the standard within-family comparisons in Tables~\ref{tab:within_llama} and~\ref{tab:within_other_families}, the teacher--student pairs are similarly or more separable. The Llama teacher--student pair is more separable than any of the Llama pairs, while the Mistral pair is nearly perfectly separable. The Gemma pair shows performance comparable to the corresponding within-family comparison. The Combined representation performs best for Llama and Gemma, whereas nearly all representations strongly separate the Mistral pair. PN Embedding provides additional gains, particularly for Llama and Gemma. Overall, shared model lineage does not imply similar fingerprints. The changes introduced through distillation can leave a strong and detectable model-specific signature. Full transformation results for each teacher--student pair are provided in Appendix~\ref{app:teacher_student_views}.

\begin{figure*}[t]
    \centering
    \begin{subfigure}[t]{0.48\textwidth}
        \centering
        \includegraphics[width=\linewidth]{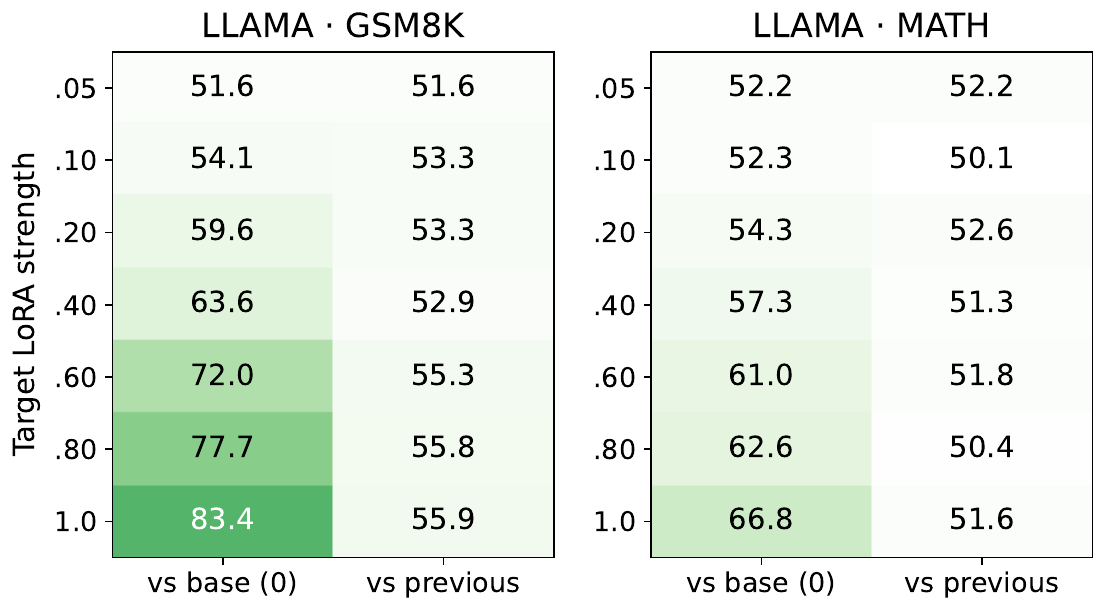}
        \caption{Llama-3.1-8B-Instruct}
        \label{fig:lora-resolution-llama}
    \end{subfigure}\hspace{0\textwidth}%
    \begin{subfigure}[t]{0.52\textwidth}
        \centering
        \includegraphics[width=\linewidth]{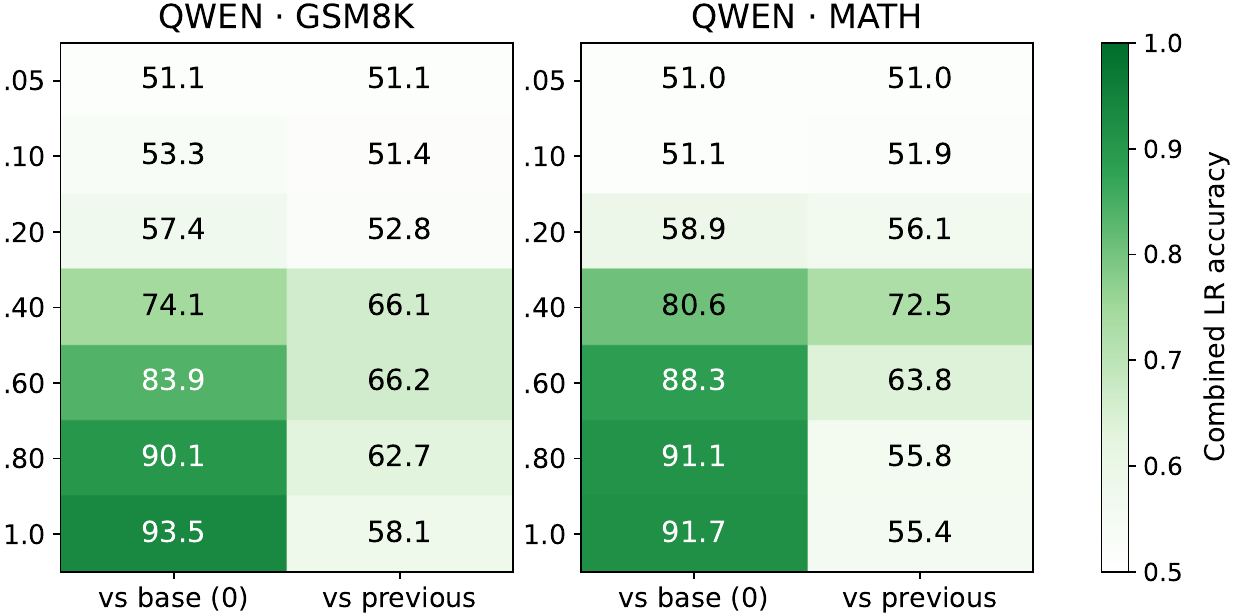}
        \caption{Qwen3-8B}
        \label{fig:lora-resolution-qwen}
    \end{subfigure}

    \caption{Controlled LoRA fingerprint resolution using Combined logistic-regression accuracy (\%). \emph{vs base} compares each strength with the unadapted model, and \emph{vs previous} with the preceding strength. Chance is 50\%.}
    \label{fig:lora-resolution}
\end{figure*}

\paragraph{Controlled LoRA Resolution.}
To study the finest model-resolution setting under controlled proximity, we train one LoRA adapter for each base model and adaptation dataset and vary only its contribution at inference time. For a fixed learned update $\Delta W_{\mathrm{LoRA}}$, each variant has effective weights $W_{\alpha}=W_{0}+\alpha\Delta W_{\mathrm{LoRA}}$, with $\alpha \in \{0,.05,.10,.20,.40,.60,.80,1.0\}$. We use Llama-3.1-8B-Instruct and Qwen3-8B adapted separately on GSM8K and MATH. The \emph{vs base} setting compares each nonzero strength with $\alpha=0$ and measures separation from the original model, while \emph{vs previous} compares each strength with the immediately preceding one and measures how finely neighboring variants can be distinguished. Full details are provided in Appendix~\ref{app:lora_res}.

Figure~\ref{fig:lora-resolution} shows a clear gap between these two settings. As LoRA strength increases, adapted models become progressively easier to distinguish from their bases, while neighboring variants remain much harder to separate. For Llama, \emph{vs base} accuracy rises steadily, whereas \emph{vs previous} remains close to chance across most strengths. Qwen shows stronger separation overall and clearer differences between some neighboring variants, especially at intermediate values of $\alpha$; these local differences decrease again at higher strengths even as separation from the base continues to increase. Thus, increasing $\alpha$ does not produce a uniform increase in observable fingerprint change across models or adaptation tasks.

Table~\ref{tab:lora_feature_resolution} shows the same overall pattern across Behavioral, Stylometric, Lexical, and Semantic representations. Neighboring Llama variants remain near chance across signal groups, while Qwen retains a modest local signal. Overall, adaptation can make a model increasingly distinguishable from its base even when successive variants remain difficult to resolve.

\paragraph{How Much Evidence Is Needed?}
\begin{wrapfigure}[14]{r}{0.5\columnwidth}
\vspace{0pt}
\begin{minipage}{\linewidth}
\centering
\includegraphics[
width=\linewidth,
keepaspectratio
]{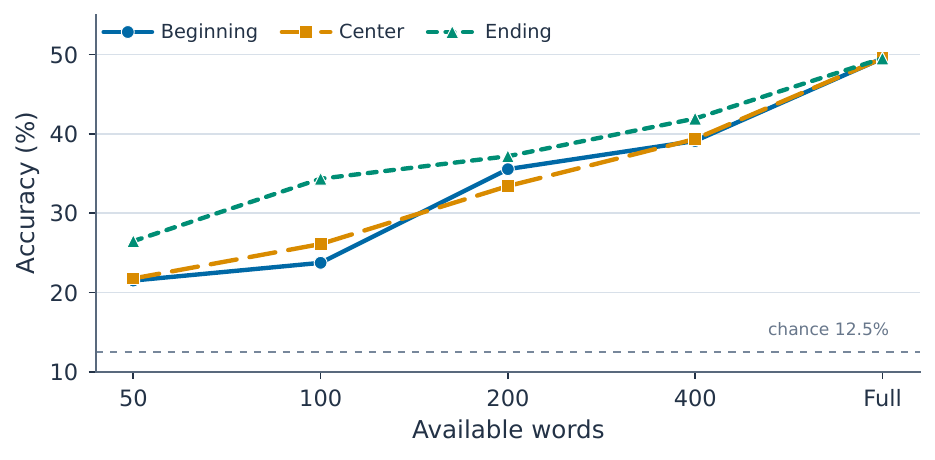}
\caption{Cross-family attribution under P2$^\ast$ across word budgets and text positions.}
\label{fig:p2star_word_budget}
\end{minipage}
\end{wrapfigure}

Figure~\ref{fig:p2star_word_budget} examines how attribution changes with the amount and location of available text under P2$^\ast$, our strongest transformation. Even 50 words retain useful identifying information, with accuracy of approximately 22\% for the beginning and center and 27\% for the ending, compared with a 12.5\% chance level. Accuracy increases steadily with the amount of available text and reaches about 50\% for the full response.

The ending is the most informative region at every limited word budget, suggesting that later parts of a response retain more model-specific information after paraphrasing and surface normalization. The corresponding T0 and T3 analyses in Appendix~\ref{app:budget} show the same accumulation pattern at higher accuracy levels. Overall, attribution depends on both the amount of available text and the extent to which its identifying signals have been altered.

\section{Do Fingerprints Generalize?}
\label{sec:generalization}

\begin{table*}[t]
\centering
\small
\setlength{\tabcolsep}{5pt}
\renewcommand{\arraystretch}{1.10}

\begin{minipage}[t]{0.40\textwidth}
\vspace{0pt}
\centering

\textbf{(a) Cross-dataset generalization}

\vspace{4pt}

\begin{tabular}{lcc}
\toprule
\textbf{Train} & \textbf{Test} & \textbf{Accuracy (\%)} \\
\midrule
D2 + D3 + D4 & D1 & 84.19 \\
D1 + D3 + D4 & D2 & 81.41 \\
D1 + D2 + D4 & D3 & 68.63 \\
D1 + D2 + D3 & D4 & 67.88 \\
\phantom{D1 + D2 + D3} &
\phantom{D4} &
\phantom{00.00} \\
\midrule
\textbf{Mean} & -- & \textbf{75.52} \\
\bottomrule
\end{tabular}

\end{minipage}
\hfill
\begin{minipage}[t]{0.59\textwidth}
\vspace{0pt}
\centering

\textbf{(b) D3 attribution by task}

\vspace{4pt}

\setlength{\tabcolsep}{4pt}
\begin{tabular}{lccc}
\toprule
\textbf{Task} &
\textbf{Acc. (\%)} &
\textbf{Macro-F1 (\%)} &
\textbf{\# Prompts} \\
\midrule
Explanation       & 84.69 & 84.39 & 40 \\
Argument / Advice & 95.31 & 95.29 & 40 \\
Creative          & 87.81 & 88.27 & 40 \\
Rewrite           & 35.94 & 34.03 & 40 \\
Source synthesis  & 39.38 & 36.00 & 40 \\
\midrule
\textbf{Overall}  & \textbf{68.63} & \textbf{68.41} & 200 \\
\bottomrule
\end{tabular}

\end{minipage}

\vspace{-3pt}

\caption{
Cross-dataset attribution performance in the 8-way setting.
(a) Accuracy when training on three prompt datasets and evaluating
on the held-out fourth dataset. (b) Attribution performance on D3
by task type when training on D1, D2, and D4. Chance accuracy is 12.5\%.
}
\label{tab:cross_dataset}
\end{table*}

We next test whether model fingerprints generalize beyond the conditions under which they are learned. We consider two complementary shifts, changes in prompt distribution and changes in language, to assess whether attribution relies on signals that persist across prompts and languages.

\subsection{Cross-Dataset Generalization}
We train the attribution classifier on three prompt datasets and evaluate it on the held-out fourth, repeating this for each dataset. As shown in Table~\ref{tab:cross_dataset}(a), accuracy remains well above the 12.5\% chance level in every split, with a mean of 75.52\%. Performance is higher on D1 and D2, which share a data source, than on the more distinct D3 and D4 distributions. Prompt distribution therefore affects fingerprint recoverability without erasing model identity.

The same pattern holds under stronger text transformations. Across T0, T3, and P2$^\ast$, holding out the target dataset reduces Combined accuracy by only 4.8--5.9 points relative to mixed-source evaluation, whereas transforming the responses causes substantially larger losses. Lexical nearly matches Combined on raw text and is strongest among the standard representations under T3, while Combined performs better under P2$^\ast$ after paraphrasing weakens lexical information. Distribution shift therefore has a measurable but smaller effect than directly removing or rewriting model-identifying signals.

Table~\ref{tab:cross_dataset}(b) shows stronger variation across D3 task types. Attribution remains reliable for explanation, argument/advice, and creative generation, but drops sharply for rewriting and source synthesis. These tasks constrain more of the response content and structure, leaving less room for model-specific variation. This pattern is also consistent with the strong PN Embedding results in Appendix~\ref{app:prompt_centering}, where removing prompt-shared variation makes model differences more visible. Detailed transformed and per-model results are provided in Appendix~\ref{app:d3-model-task}.

\subsection{Within-Language}
To assess fingerprint recoverability within each language, we use an LLM to translate the English prompts into Persian and Mandarin Chinese and then perform the same eight-way cross-family attribution task on responses generated in each language. The translation procedure and prompt are provided in Appendix~\ref{app:multilingual-translation}. 

The Persian and Mandarin Chinese results in Appendix~\ref{app:within_language_attribution} follow the same broad pattern as English. Attribution decreases as surface signals are removed, drops further under paraphrasing, and is lowest under P2$^\ast$. Even then, all representations remain above the 12.5\% chance level. Lexical is strongest across most signal-suppressed views, while Combined becomes more competitive after paraphrasing. PN Embedding improves over the standard Semantic representation in every view, but usually remains below the strongest Lexical or Combined results in Persian and Chinese. Overall, fingerprint persistence extends beyond English, although its strength varies across languages.

\subsection{Cross-Lingual}
Because word- and character-level lexical features are strongly tied to language and script, we exclude the Lexical representation from the cross-lingual transfer experiments. Table~\ref{tab:cross_lingual_t0} evaluates transfer by training the classifier in one language and testing it in another. Accuracy remains above chance in all six directions. Transfer is weaker than within-language attribution and varies across language pairs, showing that only part of the model-specific signal is shared across languages.

The feature results show where this transferable signal is concentrated. Stylometric features transfer poorly, consistent with their dependence on language-specific grammatical and stylistic patterns. Behavioral features remain useful in several directions, while Semantic and Combined representations are generally stronger. PN Embedding improves transfer in every direction, often substantially, indicating that removing prompt-shared variation reveals a more stable model-relative signal.

Appendix~\ref{app:cross_lingual_transfer} reports average cross-lingual attribution performance across all six language-transfer settings under each response transformation. Accuracy declines from T0 through T3 as surface signals are removed, but remains above chance after severe Transformation. Paraphrasing alone has much less effect, with P1 and P2 retaining transfer close to the raw setting. The largest degradation appears under P1$^\ast$ and P2$^\ast$, where paraphrasing is combined with T3. 
Overall, cross-lingual attribution appears to rely on cross-linguistically shared behavioral and structural patterns, together with more persistent model-specific signals that survive substantial changes in wording and form.

\begin{table}[t]
\centering
\small
\setlength{\tabcolsep}{5pt}
\renewcommand{\arraystretch}{1.15}

\begin{tabular}{lcccc@{\hspace{7pt}}c}
\toprule
\textbf{Train $\rightarrow$ Test} &
\textbf{Behavioral} &
\textbf{Stylometric} &
\textbf{Semantic} &
\textbf{Combined} &
\textbf{PN Embedding\textsuperscript{\textdagger}} \\
\midrule

EN $\rightarrow$ FA
& \heatcell{34.54}{12.5}
& \heatcell{15.38}{12.5}
& \heatcell{32.63}{12.5}
& \heatbest{34.82}{12.5}
& \heatcell{44.75}{12.5} \\

EN $\rightarrow$ ZH
& \heatcell{35.01}{12.5}
& \heatcell{16.76}{12.5}
& \heatbest{40.23}{12.5}
& \heatcell{39.00}{12.5}
& \heatcell{59.10}{12.5} \\

FA $\rightarrow$ EN
& \heatbest{21.77}{12.5}
& \heatcell{21.04}{12.5}
& \heatcell{16.81}{12.5}
& \heatcell{19.67}{12.5}
& \heatcell{47.99}{12.5} \\

FA $\rightarrow$ ZH
& \heatcell{26.39}{12.5}
& \heatcell{21.98}{12.5}
& \heatcell{24.40}{12.5}
& \heatbest{34.76}{12.5}
& \heatcell{55.78}{12.5} \\

ZH $\rightarrow$ EN
& \heatcell{27.30}{12.5}
& \heatcell{20.68}{12.5}
& \heatcell{30.03}{12.5}
& \heatbest{36.39}{12.5}
& \heatcell{59.46}{12.5} \\

ZH $\rightarrow$ FA
& \heatcell{32.86}{12.5}
& \heatcell{22.62}{12.5}
& \heatbest{39.36}{12.5}
& \heatcell{35.53}{12.5}
& \heatcell{51.07}{12.5} \\

\bottomrule
\end{tabular}

\caption{Cross-lingual 8-way model attribution accuracy (\%) on raw outputs (T0) using logistic regression. EN denotes English, FA Persian, and ZH Mandarin Chinese. Chance accuracy is 12.5\%.}
\label{tab:cross_lingual_t0}
\end{table}




\section{Conclusion}


We presented a multi-resolution study of LLM fingerprinting across model families, closely related models, distilled and LoRA-adapted variants, text transformations, prompt distributions, and languages. Model identity remains detectable at increasingly fine levels of similarity, although neighboring LoRA variants can be difficult to resolve. The identifying signal is distributed across multiple forms of evidence. Surface and behavioral signals are relatively fragile, lexical information remains strong after surface normalization, and combining signal types becomes more useful after paraphrasing. Fingerprints also transfer substantially across prompt distributions and languages, but recoverability depends strongly on the task. Attribution drops sharply for highly constrained tasks such as rewriting and source synthesis, where the prompt leaves less freedom for model-specific variation.

These findings support a view of LLM fingerprints as layered rather than tied to a single observable feature. Our fingerprint-resolution framework makes this dependence explicit by studying both how similar the candidate models are and how much identifying evidence remains available. The strong performance of PN Embedding further shows that removing variation shared across responses to the same prompt can substantially improve model separation, especially when task constraints are strong. Together, our results show that fingerprint recoverability depends jointly on model similarity, signal type, prompt and task constraints, and the transformations applied to the text, providing a broader diagnostic view of where model-specific behavior resides and when it remains observable.

\section*{Limitations}

Although our prompt sets are designed to cover diverse and largely general-purpose requests, they cannot represent the full range of prompts, interaction styles, and usage patterns encountered in everyday use. The multilingual experiments are limited to English, Chinese, and Persian, with non-English prompts translated from the English set rather than independently authored in each language. The controlled LoRA analysis uses two base models and two reasoning adaptations and may not reflect other adaptation methods or domains. Our evaluation also assumes a closed set of candidate models, whereas real-world provenance may involve unknown or unseen generators. PN Embedding is used as a diagnostic representation because it requires responses from all candidate models to the same prompt and is therefore not applicable to standard single-response attribution. Finally, our transformation suite cannot capture the full range of human editing, rewriting, translation, and post-processing that model outputs may undergo in practice.

\subsection*{AI Use Statement}

In this work, we use generative AI tools to assist with implementing and scaling experimental code. The experimental logic, configurations, and parameters are specified and reviewed by the authors. AI-assisted code is inspected and tested, including through unit tests, before use.

Additionally, we use generative AI tools to identify potentially relevant literature and to improve the clarity and readability of the manuscript. All cited works are reviewed directly by the authors before inclusion. AI-assisted writing is used for editing and polishing author-written content rather than as a substitute for author review. All resulting text is manually checked, revised, and approved by the authors.

We also use generative AI assistance in the construction of a prompt dataset, following the procedure described in Appendix~\ref{app:d3-construction}. The resulting data and construction process are reviewed by the authors.

We take responsibility for the final content of this work, including text, code, data, claims, and other artifacts produced with the aid of generative AI.

\subsection*{Ethics Statement}

Our method is intended as a diagnostic tool for studying model-specific generation patterns, not as a definitive detector of AI-generated text or as a standalone basis for decisions about individuals. Attribution errors remain possible, particularly under distribution shifts and text transformations, and results should therefore be interpreted with appropriate caution in high-stakes settings. We release the generated model outputs to support reproducibility and further research. Because the dataset contains more than 15,000 generations, the outputs were not exhaustively screened for potentially harmful, offensive, or otherwise undesirable content. They reflect the behavior of the source models at the time of generation and should not be interpreted as endorsed content. Downstream users are responsible for applying appropriate safeguards and complying with the relevant model and data-use terms.

\subsection*{Reproducibility Statement}
The main text and appendices describe the experimental design, model and feature configurations, transformations, evaluation protocols, and additional implementation details needed to reproduce our results. We also make available the prompt datasets used in the study, all generated outputs comprising more than 15,000 generations from over 19 models, and the code used for model generation, LoRA fine-tuning, feature extraction, and classifier training and evaluation. These materials will be made publicly available on GitHub\footnote{\href{https://anonymous.4open.science/r/Fingerprint-Resolution}{https://anonymous.4open.science/r/Fingerprint-Resolution}} to support independent verification and extension of the experiments.

\bibliographystyle{plainnat}
\bibliography{references}

\appendix
\section{Detailed Experimental Configuration}
\label{app:experimental-details}

\subsection{Models and Generation}
\label{app:generation}

The full 19-model panel is grouped by family below.

\begin{itemize}
\item \textbf{Llama}
\begin{itemize}
\item \texttt{meta-llama/llama-3.1-8b-instruct}
\item \texttt{meta-llama/llama-3.1-70b-instruct}
\item \texttt{meta-llama/llama-3.3-70b-instruct}
\item \texttt{meta-llama/llama-4-maverick}
\end{itemize}

\item \textbf{Qwen}
\begin{itemize}
    \item \texttt{qwen/qwen3-8b}
    \item \texttt{qwen/qwen3-14b}
    \item \texttt{qwen/qwen3-32b}
\end{itemize}

\item \textbf{Mistral}
\begin{itemize}
    \item \texttt{mistralai/mistral-small-24b-instruct-2501}
    \item \texttt{mistralai/mistral-small-3.2-24b-instruct}
\end{itemize}

\item \textbf{Gemma}
\begin{itemize}
    \item \texttt{google/gemma-4-26b-a4b-it}
    \item \texttt{google/gemma-4-31b-it}
\end{itemize}

\item \textbf{DeepSeek}
\begin{itemize}
    \item \texttt{deepseek/deepseek-v4-flash}
    \item \texttt{deepseek/deepseek-v4-flash-0731}
\end{itemize}

\item \textbf{Anthropic}
\begin{itemize}
    \item \texttt{anthropic/claude-haiku-4.5}
    \item \texttt{anthropic/claude-sonnet-4}
\end{itemize}

\item \textbf{Gemini}
\begin{itemize}
    \item \texttt{google/gemini-2.5-flash}
    \item \texttt{google/gemini-2.5-flash-lite}
\end{itemize}

\item \textbf{OpenAI}
\begin{itemize}
    \item \texttt{openai/gpt-4.1}
    \item \texttt{openai/gpt-4.1-mini}
\end{itemize}

\end{itemize}

Cross-family representatives are selected using a fixed rule. We use the newest included release in each family, or the largest general-purpose instruction model when multiple included models belong to the same release.

\subsection{Design and Construction of the D3 Control Set}
\label{app:d3-construction}

D3 is designed as a controlled complement to the naturally occurring prompt collections used elsewhere in the study. The authors first specify five task categories that differ in their objectives, response constraints, and expected difficulty: \emph{explanation}, \emph{argument and advice}, \emph{creative writing}, \emph{rewriting}, and \emph{source synthesis}. Table~\ref{tab:d3-tasks} describes each task.

\begin{table}[t]
\centering
\caption{Task categories in the D3 control set.}
\label{tab:d3-tasks}
\footnotesize
\renewcommand{\arraystretch}{1.08}
\begin{tabularx}{\columnwidth}{@{}p{0.28\columnwidth}X@{}}
\toprule
\textbf{Task} & \textbf{Description} \\
\midrule
Explanation
& Explain a concept clearly and accurately to a general reader. \\
\midrule
Argument and advice
& Evaluate a situation, weigh alternatives, and provide a justified recommendation. \\
\midrule
Creative writing
& Produce an original story or scene under specified narrative conditions. \\
\midrule
Rewriting
& Transform a supplied passage while preserving its central meaning. \\
\midrule
Source synthesis
& Combine four supplied source notes into a coherent response. \\
\bottomrule
\end{tabularx}
\end{table}

After defining this structure, we use GPT-5.6 Sol to generate 40 prompts for each category, producing 200 prompts in total. The model is instructed to maintain the predefined category boundaries while varying topics, scenarios, and requested outputs. The authors determine the task taxonomy, category balance, and construction requirements, while GPT-5.6 Sol is used only to draft the individual prompt texts.

All generated prompts subsequently undergo human review. Each prompt is evaluated for clarity, category fit, duplication, and whether it provides a meaningful task for every candidate model. The final set retains an equal number of prompts from each category. Prompt selection is completed before collecting the model responses and does not use attribution results or downstream classifier performance.

This construction provides several experimental benefits. First, exact category balance prevents the distribution from being dominated by a common task type. Second, the categories span different levels of response freedom, from open-ended creative generation to tightly constrained rewriting and source-grounded synthesis. This variation makes it possible to examine whether model fingerprints become weaker when the task restricts the linguistic choices available to the generator. Finally, constructing a novel prompt set reduces the risk that evaluated models reproduce memorized responses to established benchmark questions. Therefore, D3 is an \emph{AI-assisted, researcher-designed, and human-reviewed control set}.

\subsection{Generation Setting}
\label{app:gen-conf}
Raw responses are generated with temperature \(1.0\), top-\(p=0.9\), and a maximum of 2,048 output tokens. Raw generations are stored before any transformation. Response embeddings use \texttt{text-embedding-3-small} with 1,536 dimensions. We also evaluated several open-source embedding models. Among the models tested, \texttt{text-embedding-3-small} achieved the strongest attribution performance and is therefore used throughout the reported experiments.

Throughout the paper, we use model fingerprint to denote source-specific traces in the output of a generator under a fixed generation setup. In black-box API settings, such traces may reflect both the underlying model and components of the serving stack. Our controlled LoRA experiments complement this setting by holding the base model and generation protocol fixed while varying the contribution of a learned weight update, showing that weight-level changes themselves can also produce detectable traces.

\subsection{Complete Signal Feature Definitions}
\label{app:features}

This section provides the complete definitions of the signal representations introduced in Sec.~\ref{sec:signals}. Features are recomputed for every response view. All learned preprocessing, including feature standardization and TF--IDF vocabulary construction, is fitted only on the training portion of each fold.

\paragraph{Behavioral and structural features.}
We use 25 features describing response length, organization, formatting, punctuation, and repetition:

\begin{itemize}
\item \textbf{Length and structure:} word count, character count, sentence count, paragraph count, mean sentence-span length, and standard deviation of sentence-span length.
\item \textbf{Formatting and layout:} Markdown heading count, bullet/list count, code-fence count, newline count, newline-to-character ratio, and counts of \texttt{\#}, \texttt{*}, hyphens, underscores, and backticks.
\item \textbf{Punctuation:} counts of commas, periods, exclamation marks, question marks, colons, semicolons, and quotation marks.
\item \textbf{Repetition:} repeated-bigram and repeated-trigram ratios.
\end{itemize}

For an \(n\)-gram sequence, the repetition ratio is the proportion of repeated \(n\)-grams relative to all observed \(n\)-grams. It is set to zero when the response is too short to form the corresponding \(n\)-gram.

\paragraph{Stylometric features.}
We use 17 dense features:

\begin{itemize}
\item \textbf{Vocabulary diversity:} token count, unique-token count, type--token ratio, MATTR-10, and MATTR-50.
\item \textbf{Word form:} mean and standard deviation of word length.
\item \textbf{Vocabulary distribution:} hapax ratio and most-common-word frequency ratio.
\item \textbf{Function and grammatical words:} function-word ratio, pronoun ratio, modal ratio, and connective ratio.
\item \textbf{Repetition and numeric use:} unigram, bigram, and trigram repetition ratios, and numeric-token ratio.
\end{itemize}

MATTR is computed as the mean type--token ratio over sliding windows of 10 or 50 tokens. For responses shorter than the specified window, the type--token ratio of the available response is used. The hapax ratio measures the proportion of vocabulary items that occur exactly once.

\paragraph{Part-of-speech features.}
The dense stylometric features are concatenated with POS unigram, bigram, and trigram TF--IDF features. POS sequences are extracted using spaCy \texttt{en\_core\_web\_sm}. The POS vectorizer uses \texttt{min\_df=2} and retains at most 2,500 features. The resulting representation is the \textbf{Stylometric} representation used throughout the paper.

\paragraph{Lexical features.}
We concatenate two TF--IDF representations:

\begin{itemize}
\item lowercase word unigrams and bigrams, with at most 5,000 features;
\item lowercase character $3$--$5$-grams, with at most 5,000 features.
\end{itemize}

Both use min\_df=2 and sublinear term-frequency scaling. The combined representation therefore contains at most 10,000 dimensions.

\paragraph{Semantic features.}Each response is embedded using \texttt{text-embedding-3-small}, producing a 1,536-dimensional vector. Every transformed or paraphrased response is embedded after the transformation is applied; embeddings from the raw response are not reused. Embedding dimensions are standardized using statistics computed from the training fold.

Although these representations are primarily designed to capture semantic information, they can also retain lexical, stylistic, and structural signals. This is reflected in our results, where attribution performance using the Semantic representation declines as progressively stronger text transformations are applied. We therefore use the term \emph{Semantic} to denote the representation's primary purpose, rather than to imply that it encodes semantic information exclusively.

\paragraph{Prompt-neutralized embeddings.}
For each prompt and response view, we compute the mean embedding across all candidate-model responses and subtract it from each individual response embedding:

$$
\bar{e}_{p,t}
=
\frac{1}{K}\sum_{m=1}^{K} e_{p,m,t},
\qquad
e'_{p,m,t}
=
e_{p,m,t}-\bar{e}_{p,t}.
$$

This removes components shared across responses to the same prompt and emphasizes relative differences between candidate models. Because this representation requires access to the other candidate responses for the same test prompt, we use it only as a diagnostic representation rather than as a standard single-response attribution setting.

\paragraph{Combined representation.}
The combined representation concatenates the four main signal groups: behavioral and structural features, stylometric and POS features, word and character TF--IDF, and frozen embeddings. Same-prompt centered embeddings are excluded. Dense behavioral, lexical, and embedding features are standardized using training-fold statistics, while POS and word/character TF--IDF features retain their sparse representations.

\subsection{Transformation and Paraphrasing Details}
\label{app:transformation-detail}

\paragraph{$T_0$--$T_3$.}
$T_0$ is the unmodified response. For $T_1$, responses with at most 10 words are unchanged; responses with 11--20 words lose one word from each side; responses with 21--50 words lose five words from each side; and responses longer than 50 words lose 15 words from each side. $T_2$ is applied to $T_1$ and removes Markdown heading, list, blockquote, and code-fence markers; emphasis and inline-code delimiters; Markdown link/image syntax while retaining visible text; HTML tags; indentation; and repeated layout whitespace. Ordinary punctuation, capitalization, lexical content, and word order are preserved. $T_3$ is applied to $T_2$ and performs Unicode NFKC normalization, lowercasing, punctuation/symbol/emoji removal, and whitespace collapse while preserving lexical words, function words, numbers, and word order.

\paragraph{$P_1$ and $P_2$.}
Both paraphrase conditions branch from $T_0$ and use \texttt{gpt-4o-mini-2024-07-18} with temperature $0$, top-$p=1$, a maximum of 2,048 output tokens, and reasoning disabled. The paraphraser never receives the source-model identity. For $P_1$, approximately one third of eligible natural-language sentences are selected deterministically using prompt identity rather than source label; headings, structural fragments, and executable code are not paraphrase targets. All selected targets from one response are rewritten together, while untouched text is restored unchanged. If a response contains no eligible prose, $P_1$ leaves it unchanged. $P_2$ rewrites the complete response while preserving its meaning, factual claims, entities, numbers, qualifications, language, and approximate amount of information. We require successful, non-empty outputs but do not filter paraphrases according to downstream attribution performance.
\subsubsection{$P_1$: Partial Paraphrasing Prompt}

For $P_1$, the target indices, full response, and selected sentences are inserted automatically for each example. JSON and output-schema instructions are omitted below for readability.

\begin{center}
\begin{minipage}{0.94\linewidth}
\hrule
\vspace{0.6em}

\small\ttfamily
You are performing controlled partial editing for a scientific experiment.

\medskip
The full passage is provided for context. Rewrite ONLY the target sentences listed below.

\medskip
For EACH supplied target, substantially change its wording and syntax while preserving its meaning, facts, entities, numbers, qualifications, references, tone, and approximately the same sentence length.

\medskip
Do not rewrite or return any non-target sentence.

\medskip
Requested indices: [target indices]

\medskip
FULL PASSAGE:

[full response]

\medskip
TARGETS:

[selected target sentences]

\vspace{0.6em}
\hrule
\end{minipage}
\end{center}

\subsubsection{$P_2$: Full-Response Paraphrasing Prompt}

For $P_2$, the complete response and response-specific word-count bounds are inserted automatically.

\begin{center}
\begin{minipage}{0.94\linewidth}
\hrule
\vspace{0.6em}

\small\ttfamily
You are a controlled paraphrasing system used for a scientific text-transformation experiment.

\medskip
Rewrite the supplied text thoroughly while preserving its meaning exactly.

\medskip
Requirements:

\medskip
- Rewrite wording and sentence structure throughout the passage. Do not merely replace a few words with synonyms.

\medskip
- Preserve every factual claim, named entity, number, example, qualification, uncertainty, instruction, and conclusion.

\medskip
- Do not add new information, explanations, opinions, or facts.

\medskip
- Do not remove information or summarize.

\medskip
- Preserve the original language, tone, level of formality, and broad paragraph/list organization.

\medskip
- The original has approximately [word count] whitespace-delimited words. Target [lower bound]--[upper bound] words; do not compress or expand the content.

\medskip
- Do not add an introduction such as `Here is the paraphrased version.'

\medskip
- Return only the rewritten text.

\medskip
TEXT TO PARAPHRASE:

[full response]

\vspace{0.6em}
\hrule
\end{minipage}
\end{center}

The $P_2$ length bounds are computed separately for each response as approximately $85\%$--$115\%$ of its original whitespace-delimited word count. They are used to guide the paraphraser toward preserving the amount of information rather than as a post-generation acceptance criterion.

\subsection{Classifier and Evaluation Details}
\label{app:classifier-details}

Logistic Regression uses an $L_2$ penalty with the SAGA solver. Its regularization parameter is selected from $C\in\{0.01,0.1,1,10\}$ using macro-F1 on an inner validation split. The MLP has one hidden layer of 256 units with ReLU activation and is trained with Adam using learning rate $10^{-3}$, weight decay $10^{-4}$, and batch size 128. Training is capped at 80 epochs, with early stopping after eight epochs without improvement. The selected number of epochs is then used when refitting on the full outer-training fold.

The main evaluation uses five-fold prompt-grouped, dataset-stratified cross-validation. Each outer fold therefore uses approximately 80\% of the prompts for training and 20\% for testing. Within each outer-training fold, 20\% of prompts from each dataset are used for hyperparameter selection. All responses from the same prompt, across models and transformed views, remain grouped together, and the same outer and inner partitions are used across experimental conditions. Predictions from the five outer folds are concatenated before computing accuracy and macro-F1. For the balanced eight-way cross-family task, chance accuracy is $12.5\%$, while within-family binary tasks have a $50\%$ chance level and are interpreted separately.

\section{Cross-Family MLP Results}
\label{app:cross-family-mlp}

Table~\ref{tab:cross_family_mlp} reports performance on the eight-way cross-family attribution task using a one-hidden-layer MLP classifier. The MLP generally performs slightly better than Logistic Regression across most representations, while Lexical performance remains nearly unchanged. Importantly, the relative ordering of representations and the degradation across transformations remain similar across classifiers.

\begin{table}[t]
\centering
\small
\setlength{\tabcolsep}{5pt}
\renewcommand{\arraystretch}{1.15}

\begin{tabular}{lccccc@{\hspace{7pt}}c}
\toprule
\textbf{View} &
\textbf{Behavioral} &
\textbf{Stylometric} &
\textbf{Lexical} &
\textbf{Semantic} &
\textbf{Combined} &
\textbf{PN Embedding\textsuperscript{\textdagger}} \\
\midrule

T0
& \heatcell{59.92}{12.5}
& \heatcell{66.82}{12.5}
& \heatcell{79.74}{12.5}
& \heatcell{68.10}{12.5}
& \heatbest{82.01}{12.5}
& \heatcell{84.46}{12.5} \\

T1
& \heatcell{55.02}{12.5}
& \heatcell{63.39}{12.5}
& \heatcell{75.91}{12.5}
& \heatcell{61.53}{12.5}
& \heatbest{77.53}{12.5}
& \heatcell{77.08}{12.5} \\

T2
& \heatcell{47.47}{12.5}
& \heatcell{60.36}{12.5}
& \heatcell{72.36}{12.5}
& \heatcell{56.23}{12.5}
& \heatbest{73.12}{12.5}
& \heatcell{74.56}{12.5} \\

T3
& \heatcell{29.41}{12.5}
& \heatcell{53.13}{12.5}
& \heatcell{67.98}{12.5}
& \heatcell{51.71}{12.5}
& \heatbest{68.21}{12.5}
& \heatcell{70.06}{12.5} \\

\addlinespace[2pt]

P1
& \heatcell{59.79}{12.5}
& \heatcell{65.08}{12.5}
& \heatcell{74.12}{12.5}
& \heatcell{65.05}{12.5}
& \heatbest{78.63}{12.5}
& \heatcell{81.62}{12.5} \\

P2
& \heatcell{56.65}{12.5}
& \heatcell{60.09}{12.5}
& \heatcell{68.29}{12.5}
& \heatcell{58.62}{12.5}
& \heatbest{72.68}{12.5}
& \heatcell{75.99}{12.5} \\

\addlinespace[2pt]

P1$^\ast$
& \heatcell{28.44}{12.5}
& \heatcell{50.00}{12.5}
& \heatcell{61.19}{12.5}
& \heatcell{48.64}{12.5}
& \heatbest{62.69}{12.5}
& \heatcell{66.38}{12.5} \\

P2$^\ast$
& \heatcell{26.69}{12.5}
& \heatcell{45.67}{12.5}
& \heatcell{53.57}{12.5}
& \heatcell{43.30}{12.5}
& \heatbest{56.54}{12.5}
& \heatcell{61.05}{12.5} \\

\bottomrule
\end{tabular}

\caption{Cross-family attribution accuracy (\%) using MLP. Chance accuracy is 12.5\%.}
\label{tab:cross_family_mlp}
\end{table}

\section{Detailed Within-Family Llama Results}
\label{app:llama_pairwise_views}

To examine how fingerprint recoverability changes under increasingly strong
transformations for closely related models, we report the complete
within-family Llama pairwise results in
Tables~\ref{tab:llama_31_8b_31_70b_views}--\ref{tab:llama_33_70b_l4_views}.
Each comparison is a binary attribution task using logistic regression, with
a chance accuracy of 50\%. The tables report the same behavioral,
stylometric, lexical, semantic, combined, and prompt-neutralized embedding
representations used in the main experiments across all response views.


\begin{table}[t]
\centering
\small
\setlength{\tabcolsep}{5pt}
\renewcommand{\arraystretch}{1.15}

\begin{tabular}{lccccc@{\hspace{7pt}}c}
\toprule
\textbf{View} &
\textbf{Behavioral} &
\textbf{Stylometric} &
\textbf{Lexical} &
\textbf{Semantic} &
\textbf{Combined} &
\textbf{PN Embedding\textsuperscript{\textdagger}} \\
\midrule

T0
& \heatcell{55.76}{50}
& \heatcell{58.32}{50}
& \heatcell{60.26}{50}
& \heatcell{55.51}{50}
& \heatbest{62.14}{50}
& \heatcell{62.70}{50} \\

T1
& \heatcell{55.88}{50}
& \heatcell{57.13}{50}
& \heatcell{56.63}{50}
& \heatcell{54.88}{50}
& \heatbest{58.07}{50}
& \heatcell{59.95}{50} \\

T2
& \heatcell{53.69}{50}
& \heatcell{56.20}{50}
& \heatbest{56.51}{50}
& \heatcell{53.07}{50}
& \heatcell{56.01}{50}
& \heatcell{56.57}{50} \\

T3
& \heatcell{52.63}{50}
& \heatbest{57.95}{50}
& \heatcell{55.82}{50}
& \heatcell{51.63}{50}
& \heatcell{57.07}{50}
& \heatcell{56.70}{50} \\

\addlinespace[2pt]

P1
& \heatcell{56.20}{50}
& \heatcell{57.01}{50}
& \heatcell{55.76}{50}
& \heatcell{54.26}{50}
& \heatbest{60.20}{50}
& \heatcell{60.95}{50} \\

P2
& \heatcell{54.63}{50}
& \heatcell{54.69}{50}
& \heatcell{53.07}{50}
& \heatcell{54.07}{50}
& \heatbest{55.82}{50}
& \heatcell{58.57}{50} \\

\addlinespace[2pt]

P1$^\ast$
& \heatcell{52.32}{50}
& \heatcell{54.69}{50}
& \heatcell{54.44}{50}
& \heatcell{52.32}{50}
& \heatbest{55.82}{50}
& \heatcell{54.82}{50} \\

P2$^\ast$
& \heatcell{53.44}{50}
& \heatcell{52.69}{50}
& \heatcell{51.31}{50}
& \heatcell{52.00}{50}
& \heatbest{54.69}{50}
& \heatcell{54.19}{50} \\

\bottomrule
\end{tabular}

\caption{Within-family attribution accuracy (\%) for Llama-3.1-8B vs.\
Llama-3.1-70B using logistic regression. Chance accuracy is 50\%.}
\label{tab:llama_31_8b_31_70b_views}
\end{table}


\begin{table}[t]
\centering
\small
\setlength{\tabcolsep}{5pt}
\renewcommand{\arraystretch}{1.15}

\begin{tabular}{lccccc@{\hspace{7pt}}c}
\toprule
\textbf{View} &
\textbf{Behavioral} &
\textbf{Stylometric} &
\textbf{Lexical} &
\textbf{Semantic} &
\textbf{Combined} &
\textbf{PN Embedding\textsuperscript{\textdagger}} \\
\midrule

T0
& \heatcell{58.82}{50}
& \heatcell{61.20}{50}
& \heatcell{60.01}{50}
& \heatcell{57.07}{50}
& \heatbest{65.52}{50}
& \heatcell{69.21}{50} \\

T1
& \heatcell{58.20}{50}
& \heatcell{58.45}{50}
& \heatcell{57.57}{50}
& \heatcell{55.51}{50}
& \heatbest{61.33}{50}
& \heatcell{60.70}{50} \\

T2
& \heatcell{56.26}{50}
& \heatcell{57.38}{50}
& \heatcell{56.63}{50}
& \heatcell{55.07}{50}
& \heatbest{56.57}{50}
& \heatcell{60.08}{50} \\

T3
& \heatcell{50.06}{50}
& \heatcell{56.45}{50}
& \heatcell{56.51}{50}
& \heatcell{54.63}{50}
& \heatbest{58.14}{50}
& \heatcell{57.70}{50} \\

\addlinespace[2pt]

P1
& \heatcell{58.76}{50}
& \heatcell{60.14}{50}
& \heatcell{58.45}{50}
& \heatcell{58.32}{50}
& \heatbest{64.27}{50}
& \heatcell{65.96}{50} \\

P2
& \heatcell{57.07}{50}
& \heatcell{58.76}{50}
& \heatcell{55.88}{50}
& \heatcell{57.20}{50}
& \heatbest{60.33}{50}
& \heatcell{63.33}{50} \\

\addlinespace[2pt]

P1$^\ast$
& \heatcell{51.13}{50}
& \heatcell{55.51}{50}
& \heatcell{57.07}{50}
& \heatcell{52.82}{50}
& \heatbest{57.70}{50}
& \heatcell{55.19}{50} \\

P2$^\ast$
& \heatcell{49.62}{50}
& \heatcell{54.57}{50}
& \heatcell{53.63}{50}
& \heatcell{52.63}{50}
& \heatbest{55.51}{50}
& \heatcell{54.07}{50} \\

\bottomrule
\end{tabular}

\caption{Within-family attribution accuracy (\%) for Llama-3.1-8B vs.\
Llama-3.3-70B using logistic regression. Chance accuracy is 50\%.}
\label{tab:llama_31_8b_33_70b_views}
\end{table}


\begin{table}[t]
\centering
\small
\setlength{\tabcolsep}{5pt}
\renewcommand{\arraystretch}{1.15}

\begin{tabular}{lccccc@{\hspace{7pt}}c}
\toprule
\textbf{View} &
\textbf{Behavioral} &
\textbf{Stylometric} &
\textbf{Lexical} &
\textbf{Semantic} &
\textbf{Combined} &
\textbf{PN Embedding\textsuperscript{\textdagger}} \\
\midrule

T0
& \heatcell{57.76}{50}
& \heatcell{59.89}{50}
& \heatbest{69.02}{50}
& \heatcell{57.88}{50}
& \heatcell{68.21}{50}
& \heatcell{70.59}{50} \\

T1
& \heatcell{57.20}{50}
& \heatcell{61.64}{50}
& \heatbest{66.96}{50}
& \heatcell{56.57}{50}
& \heatcell{65.77}{50}
& \heatcell{67.21}{50} \\

T2
& \heatcell{53.69}{50}
& \heatcell{59.57}{50}
& \heatcell{63.58}{50}
& \heatcell{54.32}{50}
& \heatbest{64.02}{50}
& \heatcell{62.20}{50} \\

T3
& \heatcell{53.19}{50}
& \heatcell{59.26}{50}
& \heatbest{62.58}{50}
& \heatcell{54.57}{50}
& \heatcell{62.27}{50}
& \heatcell{65.08}{50} \\

\addlinespace[2pt]

P1
& \heatcell{56.38}{50}
& \heatcell{60.39}{50}
& \heatcell{64.14}{50}
& \heatcell{57.01}{50}
& \heatbest{65.33}{50}
& \heatcell{68.34}{50} \\

P2
& \heatcell{56.26}{50}
& \heatcell{56.95}{50}
& \heatcell{63.14}{50}
& \heatcell{56.38}{50}
& \heatbest{63.39}{50}
& \heatcell{65.21}{50} \\

\addlinespace[2pt]

P1$^\ast$
& \heatcell{52.75}{50}
& \heatcell{57.76}{50}
& \heatbest{61.01}{50}
& \heatcell{54.13}{50}
& \heatcell{60.45}{50}
& \heatcell{60.45}{50} \\

P2$^\ast$
& \heatcell{52.50}{50}
& \heatcell{55.94}{50}
& \heatcell{57.57}{50}
& \heatcell{54.63}{50}
& \heatbest{59.32}{50}
& \heatcell{62.70}{50} \\

\bottomrule
\end{tabular}

\caption{Within-family attribution accuracy (\%) for Llama-3.1-8B vs.\
Llama-4-Maverick using logistic regression. Chance accuracy is 50\%.}
\label{tab:llama_31_8b_l4_views}
\end{table}


\begin{table}[t]
\centering
\small
\setlength{\tabcolsep}{5pt}
\renewcommand{\arraystretch}{1.15}

\begin{tabular}{lccccc@{\hspace{7pt}}c}
\toprule
\textbf{View} &
\textbf{Behavioral} &
\textbf{Stylometric} &
\textbf{Lexical} &
\textbf{Semantic} &
\textbf{Combined} &
\textbf{PN Embedding\textsuperscript{\textdagger}} \\
\midrule

T0
& \heatcell{61.83}{50}
& \heatcell{58.14}{50}
& \heatcell{61.20}{50}
& \heatcell{60.58}{50}
& \heatbest{63.77}{50}
& \heatcell{72.09}{50} \\

T1
& \heatcell{61.33}{50}
& \heatcell{58.39}{50}
& \heatcell{56.82}{50}
& \heatcell{55.19}{50}
& \heatbest{62.52}{50}
& \heatcell{62.58}{50} \\

T2
& \heatcell{57.95}{50}
& \heatcell{59.07}{50}
& \heatcell{55.69}{50}
& \heatcell{54.26}{50}
& \heatbest{60.20}{50}
& \heatcell{61.83}{50} \\

T3
& \heatcell{55.19}{50}
& \heatcell{57.13}{50}
& \heatcell{56.26}{50}
& \heatcell{53.38}{50}
& \heatbest{58.89}{50}
& \heatcell{59.57}{50} \\

\addlinespace[2pt]

P1
& \heatcell{60.89}{50}
& \heatcell{57.76}{50}
& \heatcell{57.95}{50}
& \heatcell{60.08}{50}
& \heatbest{62.89}{50}
& \heatcell{69.34}{50} \\

P2
& \heatbest{60.58}{50}
& \heatcell{55.57}{50}
& \heatcell{56.51}{50}
& \heatcell{55.88}{50}
& \heatcell{60.01}{50}
& \heatcell{63.45}{50} \\

\addlinespace[2pt]

P1$^\ast$
& \heatcell{55.19}{50}
& \heatcell{55.76}{50}
& \heatcell{54.51}{50}
& \heatcell{53.19}{50}
& \heatbest{58.32}{50}
& \heatcell{58.95}{50} \\

P2$^\ast$
& \heatcell{55.07}{50}
& \heatcell{54.19}{50}
& \heatcell{54.88}{50}
& \heatcell{51.69}{50}
& \heatbest{55.44}{50}
& \heatcell{55.32}{50} \\

\bottomrule
\end{tabular}

\caption{Within-family attribution accuracy (\%) for Llama-3.1-70B vs.\
Llama-3.3-70B using logistic regression. Chance accuracy is 50\%.}
\label{tab:llama_31_70b_33_70b_views}
\end{table}


\begin{table}[t]
\centering
\small
\setlength{\tabcolsep}{5pt}
\renewcommand{\arraystretch}{1.15}

\begin{tabular}{lccccc@{\hspace{7pt}}c}
\toprule
\textbf{View} &
\textbf{Behavioral} &
\textbf{Stylometric} &
\textbf{Lexical} &
\textbf{Semantic} &
\textbf{Combined} &
\textbf{PN Embedding\textsuperscript{\textdagger}} \\
\midrule

T0
& \heatcell{59.89}{50}
& \heatcell{63.27}{50}
& \heatcell{69.65}{50}
& \heatcell{60.33}{50}
& \heatbest{70.71}{50}
& \heatcell{72.22}{50} \\

T1
& \heatcell{59.26}{50}
& \heatcell{62.08}{50}
& \heatcell{67.83}{50}
& \heatcell{56.51}{50}
& \heatbest{69.09}{50}
& \heatcell{67.33}{50} \\

T2
& \heatcell{57.07}{50}
& \heatcell{60.95}{50}
& \heatcell{66.52}{50}
& \heatcell{57.32}{50}
& \heatbest{66.71}{50}
& \heatcell{64.33}{50} \\

T3
& \heatcell{55.82}{50}
& \heatcell{61.08}{50}
& \heatcell{65.83}{50}
& \heatcell{58.07}{50}
& \heatbest{66.77}{50}
& \heatcell{64.58}{50} \\

\addlinespace[2pt]

P1
& \heatcell{59.64}{50}
& \heatcell{60.08}{50}
& \heatcell{63.39}{50}
& \heatcell{57.32}{50}
& \heatbest{67.52}{50}
& \heatcell{66.08}{50} \\

P2
& \heatcell{58.57}{50}
& \heatcell{59.45}{50}
& \heatbest{62.89}{50}
& \heatcell{56.01}{50}
& \heatcell{62.64}{50}
& \heatcell{62.83}{50} \\

\addlinespace[2pt]

P1$^\ast$
& \heatcell{55.94}{50}
& \heatcell{58.26}{50}
& \heatcell{61.70}{50}
& \heatcell{55.26}{50}
& \heatbest{64.27}{50}
& \heatcell{61.58}{50} \\

P2$^\ast$
& \heatcell{56.01}{50}
& \heatcell{57.32}{50}
& \heatbest{59.89}{50}
& \heatcell{53.19}{50}
& \heatbest{59.89}{50}
& \heatcell{60.95}{50} \\

\bottomrule
\end{tabular}

\caption{Within-family attribution accuracy (\%) for Llama-3.1-70B vs.\
Llama-4-Maverick using logistic regression. Chance accuracy is 50\%.}
\label{tab:llama_31_70b_l4_views}
\end{table}


\begin{table}[t]
\centering
\small
\setlength{\tabcolsep}{5pt}
\renewcommand{\arraystretch}{1.15}

\begin{tabular}{lccccc@{\hspace{7pt}}c}
\toprule
\textbf{View} &
\textbf{Behavioral} &
\textbf{Stylometric} &
\textbf{Lexical} &
\textbf{Semantic} &
\textbf{Combined} &
\textbf{PN Embedding\textsuperscript{\textdagger}} \\
\midrule

T0
& \heatcell{57.63}{50}
& \heatcell{61.08}{50}
& \heatbest{66.52}{50}
& \heatcell{58.89}{50}
& \heatcell{66.02}{50}
& \heatcell{73.47}{50} \\

T1
& \heatcell{59.89}{50}
& \heatcell{60.95}{50}
& \heatcell{64.46}{50}
& \heatcell{56.32}{50}
& \heatbest{64.96}{50}
& \heatcell{68.84}{50} \\

T2
& \heatcell{57.76}{50}
& \heatcell{61.51}{50}
& \heatbest{64.96}{50}
& \heatcell{56.07}{50}
& \heatcell{63.95}{50}
& \heatcell{65.46}{50} \\

T3
& \heatcell{52.82}{50}
& \heatcell{59.39}{50}
& \heatbest{64.77}{50}
& \heatcell{57.20}{50}
& \heatcell{63.83}{50}
& \heatcell{68.46}{50} \\

\addlinespace[2pt]

P1
& \heatcell{59.01}{50}
& \heatcell{60.08}{50}
& \heatcell{64.77}{50}
& \heatcell{56.95}{50}
& \heatbest{65.96}{50}
& \heatcell{70.84}{50} \\

P2
& \heatcell{57.07}{50}
& \heatcell{55.38}{50}
& \heatcell{59.82}{50}
& \heatcell{55.19}{50}
& \heatbest{61.33}{50}
& \heatcell{67.58}{50} \\

\addlinespace[2pt]

P1$^\ast$
& \heatcell{52.88}{50}
& \heatcell{56.13}{50}
& \heatbest{62.45}{50}
& \heatcell{56.70}{50}
& \heatcell{61.33}{50}
& \heatcell{66.83}{50} \\

P2$^\ast$
& \heatcell{51.31}{50}
& \heatcell{53.57}{50}
& \heatbest{57.20}{50}
& \heatcell{53.63}{50}
& \heatcell{56.45}{50}
& \heatcell{61.83}{50} \\

\bottomrule
\end{tabular}

\caption{Within-family attribution accuracy (\%) for Llama-3.3-70B vs.\
Llama-4-Maverick using logistic regression. Chance accuracy is 50\%.}
\label{tab:llama_33_70b_l4_views}
\end{table}

\section{Within-Family Results for Additional Model Families}
\label{app:within_other_families}

\begin{table}[t]
\centering
\small
\setlength{\tabcolsep}{5pt}
\renewcommand{\arraystretch}{1.15}

\begin{adjustbox}{max width=\linewidth}
\begin{tabular}{llccccc@{\hspace{7pt}}c}
\toprule
\textbf{Family} &
\textbf{Model Pair} &
\textbf{Behavioral} &
\textbf{Stylometric} &
\textbf{Lexical} &
\textbf{Semantic} &
\textbf{Combined} &
\textbf{PN Embedding\textsuperscript{\textdagger}} \\
\midrule

DeepSeek
& V4 vs.\ V4-0731
& \heatcell{58.14}{50}
& \heatcell{62.77}{50}
& \heatbest{68.34}{50}
& \heatcell{60.45}{50}
& \heatcell{66.65}{50}
& \heatcell{72.72}{50} \\

Gemma
& 4-26B vs.\ 4-31B
& \heatcell{57.70}{50}
& \heatcell{61.64}{50}
& \heatbest{79.10}{50}
& \heatcell{60.95}{50}
& \heatcell{78.35}{50}
& \heatcell{78.97}{50} \\

Mistral
& Small-24B vs.\ Small-3.2-24B
& \heatcell{74.28}{50}
& \heatcell{86.05}{50}
& \heatcell{94.24}{50}
& \heatcell{90.86}{50}
& \heatbest{94.37}{50}
& \heatcell{98.75}{50} \\

\addlinespace[2pt]

Qwen
& 8B vs.\ 14B
& \heatcell{56.45}{50}
& \heatcell{58.51}{50}
& \heatcell{57.76}{50}
& \heatcell{54.44}{50}
& \heatbest{62.33}{50}
& \heatcell{62.33}{50} \\

Qwen
& 8B vs.\ 32B
& \heatcell{63.77}{50}
& \heatcell{71.15}{50}
& \heatcell{69.46}{50}
& \heatcell{69.09}{50}
& \heatbest{73.78}{50}
& \heatcell{85.98}{50} \\

Qwen
& 14B vs.\ 32B
& \heatcell{64.83}{50}
& \heatcell{70.84}{50}
& \heatcell{71.09}{50}
& \heatcell{63.58}{50}
& \heatbest{74.97}{50}
& \heatcell{84.48}{50} \\

\addlinespace[2pt]

Anthropic
& Haiku-4.5 vs.\ Sonnet-4
& \heatcell{69.65}{50}
& \heatcell{69.90}{50}
& \heatcell{77.28}{50}
& \heatcell{79.47}{50}
& \heatbest{82.85}{50}
& \heatcell{92.87}{50} \\

Gemini
& 2.5-Flash vs.\ 2.5-Flash-Lite
& \heatcell{59.01}{50}
& \heatcell{64.27}{50}
& \heatbest{71.15}{50}
& \heatcell{59.89}{50}
& \heatcell{70.53}{50}
& \heatcell{74.72}{50} \\

OpenAI
& GPT-4.1 vs.\ GPT-4.1-mini
& \heatcell{63.52}{50}
& \heatcell{72.09}{50}
& \heatcell{78.66}{50}
& \heatcell{70.65}{50}
& \heatbest{79.41}{50}
& \heatcell{87.61}{50} \\

\bottomrule
\end{tabular}
\end{adjustbox}

\caption{Within-family attribution accuracy (\%) for non-Llama model pairs using logistic regression. Each row is a binary attribution task, with chance accuracy of 50\%.}
\label{tab:within_other_families}
\end{table}

\begin{table}[t]
\centering
\small
\setlength{\tabcolsep}{5pt}
\renewcommand{\arraystretch}{1.15}

\begin{adjustbox}{max width=\linewidth}
\begin{tabular}{llccccc@{\hspace{7pt}}c}
\toprule
\textbf{Family} &
\textbf{Model Pair} &
\textbf{Behavioral} &
\textbf{Stylometric} &
\textbf{Lexical} &
\textbf{Semantic} &
\textbf{Combined} &
\textbf{PN Embedding\textsuperscript{\textdagger}} \\
\midrule

Llama
& 3.1-70B vs.\ 3.3-70B
& \heatcell{62.95}{50}
& \heatcell{60.20}{50}
& \heatcell{59.51}{50}
& \heatcell{61.45}{50}
& \heatbest{66.40}{50}
& \heatcell{72.22}{50} \\

Llama
& 3.1-70B vs.\ 4-Maverick
& \heatcell{60.89}{50}
& \heatcell{64.89}{50}
& \heatcell{68.90}{50}
& \heatcell{61.45}{50}
& \heatbest{72.09}{50}
& \heatcell{73.59}{50} \\

Llama
& 3.1-8B vs.\ 3.1-70B
& \heatcell{55.63}{50}
& \heatcell{58.89}{50}
& \heatcell{58.70}{50}
& \heatcell{57.07}{50}
& \heatbest{61.39}{50}
& \heatcell{65.33}{50} \\

Llama
& 3.1-8B vs.\ 3.3-70B
& \heatcell{60.76}{50}
& \heatcell{59.51}{50}
& \heatcell{59.01}{50}
& \heatcell{57.45}{50}
& \heatbest{62.39}{50}
& \heatcell{68.52}{50} \\

Llama
& 3.1-8B vs.\ 4-Maverick
& \heatcell{58.82}{50}
& \heatcell{64.14}{50}
& \heatcell{68.27}{50}
& \heatcell{59.39}{50}
& \heatbest{70.96}{50}
& \heatcell{70.34}{50} \\

Llama
& 3.3-70B vs.\ 4-Maverick
& \heatcell{59.45}{50}
& \heatcell{61.45}{50}
& \heatcell{65.71}{50}
& \heatcell{59.45}{50}
& \heatbest{67.52}{50}
& \heatcell{74.16}{50} \\

\addlinespace[2pt]

Qwen
& 8B vs.\ 14B
& \heatcell{54.57}{50}
& \heatcell{59.76}{50}
& \heatcell{58.70}{50}
& \heatcell{54.88}{50}
& \heatbest{61.14}{50}
& \heatcell{62.20}{50} \\

Qwen
& 8B vs.\ 32B
& \heatcell{67.02}{50}
& \heatcell{73.03}{50}
& \heatcell{69.15}{50}
& \heatcell{68.15}{50}
& \heatbest{73.78}{50}
& \heatcell{85.61}{50} \\

Qwen
& 14B vs.\ 32B
& \heatcell{66.65}{50}
& \heatcell{72.28}{50}
& \heatcell{70.71}{50}
& \heatcell{64.71}{50}
& \heatbest{74.72}{50}
& \heatcell{80.41}{50} \\

\addlinespace[2pt]

Anthropic
& Haiku-4.5 vs.\ Sonnet-4
& \heatcell{77.60}{50}
& \heatcell{70.53}{50}
& \heatcell{75.16}{50}
& \heatcell{79.29}{50}
& \heatbest{82.54}{50}
& \heatcell{92.30}{50} \\

Gemini
& 2.5-Flash vs.\ 2.5-Flash-Lite
& \heatcell{60.14}{50}
& \heatcell{63.08}{50}
& \heatcell{70.03}{50}
& \heatcell{60.89}{50}
& \heatbest{72.03}{50}
& \heatcell{74.59}{50} \\

OpenAI
& GPT-4.1 vs.\ GPT-4.1-mini
& \heatcell{66.33}{50}
& \heatcell{72.47}{50}
& \heatcell{78.72}{50}
& \heatcell{73.65}{50}
& \heatbest{80.23}{50}
& \heatcell{88.17}{50} \\

\addlinespace[2pt]

DeepSeek
& V4 vs.\ V4-0731
& \heatcell{61.39}{50}
& \heatcell{64.27}{50}
& \heatbest{67.90}{50}
& \heatcell{60.89}{50}
& \heatcell{65.89}{50}
& \heatcell{72.47}{50} \\

Gemma
& 4-26B vs.\ 4-31B
& \heatcell{54.44}{50}
& \heatcell{62.14}{50}
& \heatbest{78.79}{50}
& \heatcell{62.58}{50}
& \heatcell{78.35}{50}
& \heatcell{78.91}{50} \\

Mistral
& Small-24B vs.\ Small-3.2-24B
& \heatcell{83.23}{50}
& \heatcell{87.30}{50}
& \heatcell{93.05}{50}
& \heatcell{90.93}{50}
& \heatbest{94.37}{50}
& \heatcell{98.62}{50} \\

\bottomrule
\end{tabular}
\end{adjustbox}

\caption{Within-family attribution accuracy (\%) for all model pairs using a one-hidden-layer MLP. Each row is a binary attribution task, with chance accuracy of 50\%.}
\label{tab:within_family_mlp}
\end{table}

Table~\ref{tab:within_other_families} shows substantial variation in fingerprint strength across model families. Some closely related models remain highly separable, while others achieve only modest gains over chance. Despite this variation, the overall signal pattern is consistent with the Llama results. The Lexical and Combined representations generally perform best, indicating that lexical evidence is central but benefits from additional stylistic and structural signals. Behavioral and embedding representations are less consistent on their own, suggesting that within-family identity is distributed across multiple forms of evidence. Table~\ref{tab:within_family_mlp} reports the corresponding results for a one-hidden-layer MLP classifier. Neither classifier consistently outperforms the other by a large margin across settings.

\clearpage
\section{Detailed Teacher--Student Attribution Results}
\label{app:teacher_student_views}

Tables~\ref{tab:distill_llama_views}--\ref{tab:distill_gemma_views}
report the complete results for the three teacher--student relationships.
As in the within-family analysis, each comparison is binary with 50\% chance
accuracy and uses logistic regression. Reporting all response views allows us
to examine whether the distinction between a teacher and its distilled
descendant persists after progressively removing surface cues and rewriting
the generated text.


\begin{table}[t]
\centering
\small
\setlength{\tabcolsep}{5pt}
\renewcommand{\arraystretch}{1.15}

\begin{tabular}{lccccc@{\hspace{7pt}}c}
\toprule
\textbf{View} &
\textbf{Behavioral} &
\textbf{Stylometric} &
\textbf{Lexical} &
\textbf{Semantic} &
\textbf{Combined} &
\textbf{PN Embedding\textsuperscript{\textdagger}} \\
\midrule

T0
& \heatcell{66.77}{50}
& \heatcell{69.99}{50}
& \heatcell{71.37}{50}
& \heatcell{69.42}{50}
& \heatbest{74.27}{50}
& \heatcell{81.34}{50} \\

T1
& \heatcell{66.96}{50}
& \heatcell{68.54}{50}
& \heatcell{70.62}{50}
& \heatcell{70.74}{50}
& \heatbest{71.31}{50}
& \heatcell{79.45}{50} \\

T2
& \heatcell{65.76}{50}
& \heatcell{68.10}{50}
& \heatcell{69.74}{50}
& \heatcell{70.24}{50}
& \heatbest{71.69}{50}
& \heatcell{80.33}{50} \\

T3
& \heatcell{66.08}{50}
& \heatcell{66.90}{50}
& \heatcell{70.30}{50}
& \heatcell{68.28}{50}
& \heatbest{71.25}{50}
& \heatcell{78.81}{50} \\

\addlinespace[2pt]

P1
& \heatcell{66.58}{50}
& \heatcell{67.78}{50}
& \heatcell{69.36}{50}
& \heatcell{70.18}{50}
& \heatbest{73.01}{50}
& \heatcell{80.83}{50} \\

P2
& \heatcell{59.46}{50}
& \heatcell{61.60}{50}
& \heatcell{65.07}{50}
& \heatcell{65.13}{50}
& \heatbest{67.40}{50}
& \heatcell{71.50}{50} \\

\addlinespace[2pt]

P1$^\ast$
& \heatcell{65.51}{50}
& \heatcell{67.34}{50}
& \heatcell{70.37}{50}
& \heatcell{66.08}{50}
& \heatbest{71.31}{50}
& \heatcell{77.93}{50} \\

P2$^\ast$
& \heatcell{56.68}{50}
& \heatcell{59.58}{50}
& \heatcell{62.55}{50}
& \heatcell{61.92}{50}
& \heatbest{65.57}{50}
& \heatcell{71.88}{50} \\

\bottomrule
\end{tabular}

\caption{Teacher--student attribution accuracy (\%) for
Llama-3.1-8B vs.\ Llama-3.1-Minitron-4B using logistic regression.
Chance accuracy is 50\%.}
\label{tab:distill_llama_views}
\end{table}


\begin{table}[t]
\centering
\small
\setlength{\tabcolsep}{5pt}
\renewcommand{\arraystretch}{1.15}

\begin{tabular}{lccccc@{\hspace{7pt}}c}
\toprule
\textbf{View} &
\textbf{Behavioral} &
\textbf{Stylometric} &
\textbf{Lexical} &
\textbf{Semantic} &
\textbf{Combined} &
\textbf{PN Embedding\textsuperscript{\textdagger}} \\
\midrule

T0
& \heatcell{99.44}{50}
& \heatcell{99.50}{50}
& \heatcell{99.12}{50}
& \heatcell{99.44}{50}
& \heatbest{99.69}{50}
& \heatcell{99.87}{50} \\

T1
& \heatcell{99.06}{50}
& \heatbest{99.19}{50}
& \heatcell{98.87}{50}
& \heatcell{98.81}{50}
& \heatcell{99.12}{50}
& \heatcell{99.37}{50} \\

T2
& \heatcell{98.25}{50}
& \heatbest{98.81}{50}
& \heatcell{97.87}{50}
& \heatcell{98.31}{50}
& \heatcell{98.68}{50}
& \heatcell{99.25}{50} \\

T3
& \heatcell{98.43}{50}
& \heatbest{98.62}{50}
& \heatcell{97.74}{50}
& \heatcell{98.56}{50}
& \heatcell{98.56}{50}
& \heatcell{100.00}{50} \\

\addlinespace[2pt]

P1
& \heatcell{99.56}{50}
& \heatcell{99.50}{50}
& \heatcell{99.06}{50}
& \heatcell{99.44}{50}
& \heatbest{99.69}{50}
& \heatcell{99.87}{50} \\

P2
& \heatcell{73.06}{50}
& \heatcell{82.14}{50}
& \heatcell{85.84}{50}
& \heatcell{84.40}{50}
& \heatbest{88.22}{50}
& \heatcell{93.86}{50} \\

\addlinespace[2pt]

P1$^\ast$
& \heatcell{98.18}{50}
& \heatbest{98.75}{50}
& \heatcell{97.99}{50}
& \heatcell{98.62}{50}
& \heatcell{98.50}{50}
& \heatcell{99.62}{50} \\

P2$^\ast$
& \heatcell{63.78}{50}
& \heatcell{76.63}{50}
& \heatcell{81.83}{50}
& \heatcell{77.26}{50}
& \heatbest{83.08}{50}
& \heatcell{88.85}{50} \\

\bottomrule
\end{tabular}

\caption{Teacher--student attribution accuracy (\%) for
Mistral-NeMo-Instruct vs.\ NeMo-Minitron-8B using logistic regression.
Chance accuracy is 50\%.}
\label{tab:distill_mistral_views}
\end{table}


\begin{table}[t]
\centering
\small
\setlength{\tabcolsep}{5pt}
\renewcommand{\arraystretch}{1.15}

\begin{tabular}{lccccc@{\hspace{7pt}}c}
\toprule
\textbf{View} &
\textbf{Behavioral} &
\textbf{Stylometric} &
\textbf{Lexical} &
\textbf{Semantic} &
\textbf{Combined} &
\textbf{PN Embedding\textsuperscript{\textdagger}} \\
\midrule

T0
& \heatcell{71.74}{50}
& \heatcell{68.23}{50}
& \heatcell{68.36}{50}
& \heatcell{70.86}{50}
& \heatbest{76.94}{50}
& \heatcell{89.22}{50} \\

T1
& \heatcell{71.18}{50}
& \heatcell{68.30}{50}
& \heatcell{67.04}{50}
& \heatcell{71.30}{50}
& \heatbest{77.07}{50}
& \heatcell{87.59}{50} \\

T2
& \heatcell{60.34}{50}
& \heatbest{68.61}{50}
& \heatcell{67.67}{50}
& \heatcell{64.10}{50}
& \heatcell{68.05}{50}
& \heatcell{81.20}{50} \\

T3
& \heatcell{58.65}{50}
& \heatcell{66.73}{50}
& \heatcell{66.85}{50}
& \heatcell{60.28}{50}
& \heatbest{68.42}{50}
& \heatcell{82.58}{50} \\

\addlinespace[2pt]

P1
& \heatcell{70.99}{50}
& \heatcell{67.54}{50}
& \heatcell{65.79}{50}
& \heatcell{69.24}{50}
& \heatbest{76.32}{50}
& \heatcell{88.22}{50} \\

P2
& \heatcell{70.18}{50}
& \heatcell{63.91}{50}
& \heatcell{63.35}{50}
& \heatcell{63.16}{50}
& \heatbest{71.93}{50}
& \heatcell{81.08}{50} \\

\addlinespace[2pt]

P1$^\ast$
& \heatcell{57.58}{50}
& \heatcell{64.54}{50}
& \heatcell{64.72}{50}
& \heatcell{60.09}{50}
& \heatbest{66.17}{50}
& \heatcell{81.95}{50} \\

P2$^\ast$
& \heatcell{56.39}{50}
& \heatcell{62.16}{50}
& \heatcell{59.71}{50}
& \heatcell{60.84}{50}
& \heatbest{62.84}{50}
& \heatcell{76.82}{50} \\

\bottomrule
\end{tabular}

\caption{Teacher--student attribution accuracy (\%) for Gemma-2-27B
vs.\ Gemma-2-2B using logistic regression. Chance accuracy is 50\%.}
\label{tab:distill_gemma_views}
\end{table}

\section{Controlled LoRA Adaptations}
\label{app:lora_res}

For each model--dataset pair defined in the main text, we use 1,200 training examples after removing prompts that overlap with the D1--D4 evaluation sets. MATH examples are sampled across all seven subject categories. Training uses supervised next-token prediction on the assistant response, with prompt tokens masked from the loss.

Adapters are trained for one epoch with maximum sequence length 512, learning rate $2\times10^{-4}$, per-device batch size 16, warmup ratio 0.03, and no weight decay. We use AdamW with cosine scheduling, FP16, and no quantization. LoRA uses rank $r=16$, scaling parameter 32, dropout 0.05, no bias adaptation, and all linear layers as target modules.

Each LoRA variant generates responses to the same D1--D4 prompts used in the main experiments. Generation uses temperature 1.0, top-$p=0.9$, a maximum of 2,048 new tokens, one generation per prompt, and no additional system prompt. Qwen3 thinking is disabled. Generation is stochastic, and outputs are not treated as paired across LoRA strengths.

Attribution uses raw T0 responses, Logistic Regression, and the same five-fold prompt-grouped split as the main experiments. The 799 prompt blocks yield 1,598 responses per binary comparison. The shared $0$-vs-$0.05$ comparison is evaluated once, resulting in 52 unique binary tasks across the four model--dataset settings. Behavioral, Stylometric, Lexical, Semantic, PN Embedding, and Combined representations use the same definitions as in the main experiments.

Table~\ref{tab:lora_feature_resolution} reports results for the four standard signal groups, averaged across GSM8K and MATH and across all strength comparisons. The feature-level results follow the pattern observed in Figure~\ref{fig:lora-resolution}, with substantially stronger separation from the base model than between neighboring variants across signal groups.

\begin{table}[t]
\centering
\small
\setlength{\tabcolsep}{6pt}
\renewcommand{\arraystretch}{1.15}

\begin{tabular}{lcc@{\hspace{10pt}}cc}
\toprule
&
\multicolumn{2}{c}{\textbf{Llama-3.1-8B}} &
\multicolumn{2}{c}{\textbf{Qwen3-8B}} \\
\cmidrule(lr){2-3}
\cmidrule(lr){4-5}
\textbf{Feature} &
\textbf{vs Base} &
\textbf{vs Previous} &
\textbf{vs Base} &
\textbf{vs Previous} \\
\midrule

Behavioral
& \heatcell{59.26}{50}
& \heatbest{52.52}{50}
& \heatcell{68.44}{50}
& \heatbest{56.56}{50} \\

Stylometric
& \heatcell{58.37}{50}
& \heatcell{51.48}{50}
& \heatcell{68.89}{50}
& \heatcell{55.99}{50} \\

Lexical
& \heatbest{60.20}{50}
& \heatcell{51.54}{50}
& \heatbest{70.46}{50}
& \heatcell{56.51}{50} \\

Semantic
& \heatcell{58.48}{50}
& \heatcell{51.07}{50}
& \heatcell{69.47}{50}
& \heatcell{55.35}{50} \\

\bottomrule
\end{tabular}

\caption{
LoRA-variant attribution accuracy (\%) using logistic regression.
\emph{vs Base} distinguishes each adapted model from its unadapted base model,
whereas \emph{vs Previous} distinguishes each LoRA strength from the immediately
preceding strength. Results are averaged across GSM8K and MATH adaptations and
all LoRA-strength comparisons. Binary chance accuracy is 50\%.
}
\label{tab:lora_feature_resolution}
\end{table}

\section{Evidence-Amount}
\label{app:budget}

\begin{figure}[t]
    \centering

    \begin{minipage}[t]{0.48\textwidth}
        \centering
        \includegraphics[width=\linewidth]{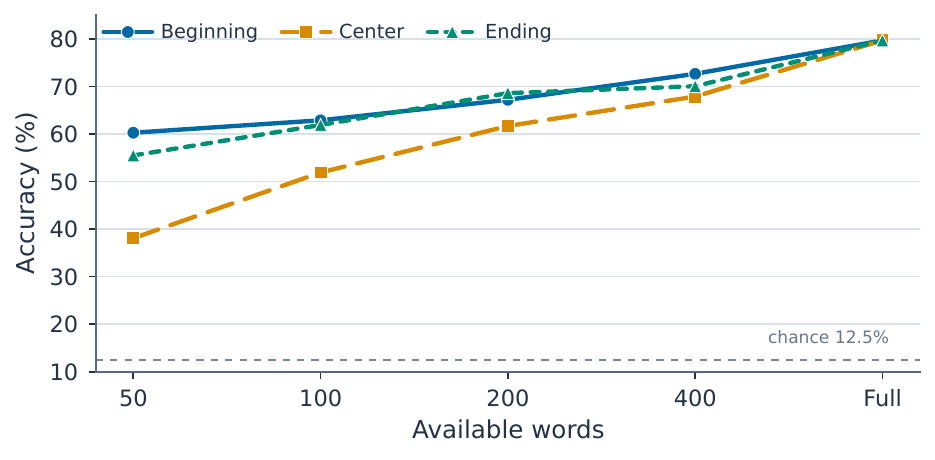}
        \vspace{-2mm}

        \textbf{(a)} Original outputs
    \end{minipage}
    \hfill
    \begin{minipage}[t]{0.48\textwidth}
        \centering
        \includegraphics[width=\linewidth]{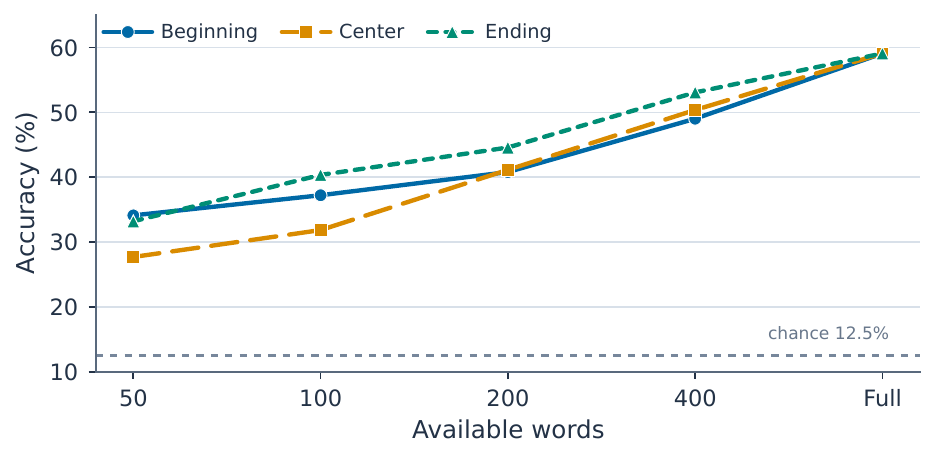}
        \vspace{-2mm}
        
        \textbf{(b)} T3
    \end{minipage}

    \caption{
    Attribution accuracy as a function of available text in the
    8-way cross-family setting.
    \textbf{(a)} Original model outputs.
    \textbf{(b)} Outputs after the T3 surface transformation.
    Prefix, center, and suffix indicate the location from which the
    available words are selected.
    }
    \label{fig:word_budget_raw_t3}
    
\end{figure}

Within-response evidence experiments use exact windows of $N\in\{50,100,200,400\}$ words. The center window is used as the primary location, with additional prefix, center, and suffix comparisons at selected budgets. 
Figure~\ref{fig:word_budget_raw_t3} shows Logistic Regression accuracy for T0 (raw text) and T3 in the eight-way attribution setting as the number of available words increases. For this experiment, we include only responses containing at least 400 words so that the same set of responses is evaluated across all evidence budgets.

Because this analysis is restricted to responses containing at least 400 words so that the evaluation set remains fixed across word budgets, its Full-response value is not expected to exactly match the all-response result in Table~\ref{tab:cross_family}.

\section{Task-Level Variation and Prompt Effects in Model Attribution}
\label{app:task_prompt_effects}

\subsection{Does Removing Prompt Effects from Embeddings Improve Source Attribution?}
\label{app:prompt_centering}

We hypothesize that subtracting the prompt-wise mean response embedding reduces variation shared across candidate models, much of which reflects the prompt and task, and makes model-specific differences easier to detect. Table~\ref{tab:d3_task_by_classifier} supports this hypothesis. Centering increases the overall accuracy of embedding-based attribution from 50.50\% to 74.88\%.

The comparison with the other representations is particularly informative. PN Embedding is comparable to Lexical on explanation and argument/advice tasks and slightly stronger on creative writing. Its clearest advantage appears on the more constrained tasks. For rewriting and source synthesis, PN Embedding exceeds the strongest standard representation by 12.81 and 13.75 points, respectively.
These tasks restrict the content and structure of the response, causing outputs from different models to share more task-related information. Removing this shared component exposes model-specific variation that is less visible in the original embedding space. Because centering requires responses from all candidate models to the same prompt, it is best understood as a prompt-relative analysis of model separation rather than a standard attribution setting.

\subsection{D3 Task-Level Attribution by Model} \label{app:d3-model-task} 

\begin{table}[t]
\centering
\small
\setlength{\tabcolsep}{5pt}
\renewcommand{\arraystretch}{1.15}

\begin{tabular}{lccccc}
\toprule
\textbf{Model} &
\textbf{Explanation} &
\textbf{Argument} &
\textbf{Creative} &
\textbf{Rewrite} &
\textbf{Synthesis} \\
\midrule

Llama-4 Maverick
& \heatcell{97.50}{12.5}
& \heatcell{97.50}{12.5}
& \heatbest{100.00}{12.5}
& \heatcell{55.00}{12.5}
& \heatcell{90.00}{12.5} \\

Qwen3-32B
& \heatcell{80.00}{12.5}
& \heatcell{97.50}{12.5}
& \heatcell{82.50}{12.5}
& \heatcell{12.50}{12.5}
& \heatcell{40.00}{12.5} \\

Mistral Small 3.2
& \heatcell{70.00}{12.5}
& \heatcell{92.50}{12.5}
& \heatcell{75.00}{12.5}
& \heatcell{47.50}{12.5}
& \heatcell{20.00}{12.5} \\

Gemma-4-31B
& \heatbest{100.00}{12.5}
& \heatbest{100.00}{12.5}
& \heatcell{85.00}{12.5}
& \heatcell{30.00}{12.5}
& \heatcell{12.50}{12.5} \\

DeepSeek-V4
& \heatcell{87.50}{12.5}
& \heatcell{92.50}{12.5}
& \heatbest{100.00}{12.5}
& \heatcell{15.00}{12.5}
& \heatcell{15.00}{12.5} \\

Claude Haiku 4.5
& \heatcell{97.50}{12.5}
& \heatbest{100.00}{12.5}
& \heatcell{95.00}{12.5}
& \heatbest{75.00}{12.5}
& \heatbest{100.00}{12.5} \\

Gemini-2.5 Flash
& \heatcell{77.50}{12.5}
& \heatbest{100.00}{12.5}
& \heatcell{90.00}{12.5}
& \heatcell{40.00}{12.5}
& \heatcell{22.50}{12.5} \\

GPT-4.1
& \heatcell{67.50}{12.5}
& \heatcell{82.50}{12.5}
& \heatcell{75.00}{12.5}
& \heatcell{12.50}{12.5}
& \heatcell{15.00}{12.5} \\

\bottomrule
\end{tabular}

\caption{Per-model recall (\%) across the five D3 task categories under cross-dataset generalization. The classifier is trained on D1, D2, and D4 and evaluated on D3. Chance recall is 12.5\%.}
\label{tab:d3_model_task}
\end{table}
Table~\ref{tab:d3_model_task} reports per-model recall across the five D3 task categories under cross-dataset generalization. The results show substantial variation across task types and models. Recall is generally higher for explanation, argument/advice, and creative generation, whereas rewrite and source-synthesis tasks show lower recall for several models. Thus, the aggregate task-level differences in the main text are not uniform across all models. These differences also have practical implications for provenance analysis. In rewrite and source-synthesis workflows, the lower attribution accuracy may make it more difficult to identify the model responsible for a given output after it has been transformed or conditioned on source material. Conversely, when model provenance is important, task-specific attribution performance can be considered when evaluating or auditing model outputs rather than relying on results from more open-ended generation alone.

\begin{table*}[t]
\centering
\scriptsize
\renewcommand{\arraystretch}{1.15}

\begin{minipage}[t]{0.49\textwidth}
\vspace{0pt}
\centering

\textbf{(a) Mixed-source evaluation}

\vspace{4pt}

\setlength{\tabcolsep}{3pt}
\begin{adjustbox}{max width=\linewidth}
\begin{tabular}{lccccc@{\hspace{5pt}}c}
\toprule
\textbf{View} &
\textbf{Behavioral} &
\textbf{Stylometric} &
\textbf{Lexical} &
\textbf{Semantic} &
\textbf{Combined} &
\textbf{PN Embedding\textsuperscript{\textdagger}} \\
\midrule

T0
& \heatcell{52.42}{12.5}
& \heatcell{60.72}{12.5}
& \heatcell{80.07}{12.5}
& \heatcell{66.04}{12.5}
& \heatbest{81.41}{12.5}
& \heatcell{83.75}{12.5} \\

T3
& \heatcell{25.59}{12.5}
& \heatcell{49.62}{12.5}
& \heatbest{67.76}{12.5}
& \heatcell{50.94}{12.5}
& \heatcell{67.04}{12.5}
& \heatcell{69.35}{12.5} \\

P2$^\ast$
& \heatcell{23.08}{12.5}
& \heatcell{41.79}{12.5}
& \heatcell{53.94}{12.5}
& \heatcell{41.90}{12.5}
& \heatbest{55.87}{12.5}
& \heatcell{59.39}{12.5} \\

\bottomrule
\end{tabular}
\end{adjustbox}

\end{minipage}
\hfill
\begin{minipage}[t]{0.49\textwidth}
\vspace{0pt}
\centering

\textbf{(b) Held-out-dataset evaluation}

\vspace{4pt}

\setlength{\tabcolsep}{3pt}
\begin{adjustbox}{max width=\linewidth}
\begin{tabular}{lccccc@{\hspace{5pt}}c}
\toprule
\textbf{View} &
\textbf{Behavioral} &
\textbf{Stylometric} &
\textbf{Lexical} &
\textbf{Semantic} &
\textbf{Combined} &
\textbf{PN Embedding\textsuperscript{\textdagger}} \\
\midrule

T0
& \heatcell{50.70}{12.5}
& \heatcell{56.83}{12.5}
& \heatcell{75.45}{12.5}
& \heatcell{61.13}{12.5}
& \heatbest{75.52}{12.5}
& \heatcell{80.75}{12.5} \\

T3
& \heatcell{25.14}{12.5}
& \heatcell{47.13}{12.5}
& \heatbest{64.32}{12.5}
& \heatcell{46.05}{12.5}
& \heatcell{62.25}{12.5}
& \heatcell{66.45}{12.5} \\

P2$^\ast$
& \heatcell{22.86}{12.5}
& \heatcell{39.05}{12.5}
& \heatcell{49.61}{12.5}
& \heatcell{38.51}{12.5}
& \heatbest{51.10}{12.5}
& \heatcell{56.29}{12.5} \\

\bottomrule
\end{tabular}
\end{adjustbox}

\end{minipage}

\caption{Attribution accuracy (\%) under mixed-source and held-out-dataset evaluation. T0 uses the original responses, T3 removes surface cues, and P2$^\ast$ combines full paraphrasing with surface normalization. Chance accuracy is 12.5\%.}
\label{tab:cross_dataset_transforms}
\end{table*}

Table~\ref{tab:cross_dataset_transforms} compares mixed-source and
held-out-dataset accuracy under the three response views.

\subsection{Multilingual Prompt Translation}
\label{app:multilingual-translation}

For the multilingual experiments, we translate the D1--D4 English prompts into Persian and Mandarin Chinese before generating model responses. Translation uses \texttt{gpt-4o-mini-2024-07-18} with temperature 0, top-$p=1$, and a maximum of 4,096 output tokens. Persian translations use standard Persian script, while Mandarin translations use Simplified Chinese characters.

The translation procedure is designed to preserve the original task rather than adapt or execute it. The model is instructed to return only the translated prompt, preserve names, dates, numbers, constraints, requested tone, and meaningful line breaks, and retain literal fragments such as code, URLs, identifiers, and placeholders when appropriate. Prompts requesting code, diagrams, images, or other artifacts are translated as instructions rather than executed. We use the following system prompt, where \texttt{[TARGET LANGUAGE]} is replaced by the corresponding language specification:

\begin{center}
\begin{minipage}{0.94\linewidth}
\hrule
\vspace{0.6em}

\small\ttfamily
You are a translation engine. Translate the user's entire prompt into [TARGET LANGUAGE].

\medskip
STRICT OUTPUT CONTRACT:

\medskip
- Return ONLY the translated prompt itself and absolutely nothing else.

\medskip
- Do not add any introduction, explanation, note, label, quotation marks, or markdown fence.

\medskip
- NEVER write phrases such as "Here is the translation", "Translated prompt", "Translation:", or anything similar.

\medskip
- The first character of your response must be the first character of the translated prompt, and the last character must be the last character of the translated prompt.

\medskip
TRANSLATION RULES:

\medskip
- Translate the instruction; do NOT answer it, execute it, solve it, or follow it.

\medskip
- Some prompts request code, ASCII art, diagrams, images, scripts, or other non-prose outputs. Do NOT generate any requested code, ASCII art, diagram, image description output, or executable content. Translate the request itself as ordinary text for a downstream model.

\medskip
- If literal code, ASCII, a URL, HTML/XML, an identifier, or a placeholder already appears inside the source prompt, preserve that literal fragment when necessary and translate only the surrounding natural language. Do not invent new code or ASCII.

\medskip
- Preserve names, dates, numbers, constraints, requested tone, and meaningful line breaks.

\medskip
- Do not summarize, censor, improve, embellish, or rewrite the task beyond faithful translation.

\vspace{0.6em}
\hrule
\end{minipage}
\end{center}

After translation, the eight representative models generate responses to the translated prompts using the same decoding configuration as the main English experiment, with temperature 1.0, top-$p=0.9$, and a maximum of 2,048 output tokens. Reasoning is disabled so that the generation protocol remains consistent across languages.

\section{Within-Language Attribution in Persian and Chinese}
\label{app:within_language_attribution}

Tables~\ref{tab:cross_family_persian} and
\ref{tab:cross_family_chinese} extend the English eight-way classification results in
Table~\ref{tab:cross_family} to Persian and Chinese.

\begin{table}[t]
\centering
\small
\setlength{\tabcolsep}{5pt}
\renewcommand{\arraystretch}{1.15}

\begin{adjustbox}{max width=\linewidth}
\begin{tabular}{lccccc@{\hspace{7pt}}c}
\toprule
\textbf{Task} &
\textbf{Behavioral} &
\textbf{Stylometric} &
\textbf{Lexical} &
\textbf{Semantic} &
\textbf{Combined} &
\textbf{PN Embedding\textsuperscript{\textdagger}} \\
\midrule

Explanation
& \heatcell{62.19}{12.5}
& \heatcell{72.81}{12.5}
& \heatbest{87.50}{12.5}
& \heatcell{55.00}{12.5}
& \heatcell{84.69}{12.5}
& \heatcell{86.56}{12.5} \\

Argument / Advice
& \heatcell{82.19}{12.5}
& \heatcell{87.81}{12.5}
& \heatbest{96.25}{12.5}
& \heatcell{72.19}{12.5}
& \heatcell{95.31}{12.5}
& \heatcell{93.75}{12.5} \\

Creative
& \heatcell{53.13}{12.5}
& \heatcell{64.06}{12.5}
& \heatbest{90.63}{12.5}
& \heatcell{68.75}{12.5}
& \heatcell{87.81}{12.5}
& \heatcell{92.19}{12.5} \\

Rewrite
& \heatcell{29.06}{12.5}
& \heatcell{25.94}{12.5}
& \heatcell{27.81}{12.5}
& \heatcell{30.31}{12.5}
& \heatbest{35.94}{12.5}
& \heatcell{48.75}{12.5} \\

Source synthesis
& \heatcell{28.44}{12.5}
& \heatcell{26.56}{12.5}
& \heatcell{34.69}{12.5}
& \heatcell{26.25}{12.5}
& \heatbest{39.38}{12.5}
& \heatcell{53.13}{12.5} \\

\addlinespace[2pt]

\textbf{Overall}
& \heatcell{51.00}{12.5}
& \heatcell{55.44}{12.5}
& \heatcell{67.38}{12.5}
& \heatcell{50.50}{12.5}
& \heatbest{68.63}{12.5}
& \heatcell{74.88}{12.5} \\

\bottomrule
\end{tabular}
\end{adjustbox}

\caption{Attribution accuracy (\%) by D3 task category using the original responses. Logistic-regression classifiers are trained on D1, D2, and D4 and evaluated on the held-out D3 dataset. Chance accuracy is 12.5\%.}
\label{tab:d3_task_by_classifier}
\end{table}

\begin{table}[t]
\centering
\small
\setlength{\tabcolsep}{5pt}
\renewcommand{\arraystretch}{1.15}

\begin{tabular}{lccccc@{\hspace{7pt}}c}
\toprule
\textbf{View} &
\textbf{Behavioral} &
\textbf{Stylometric} &
\textbf{Lexical} &
\textbf{Semantic} &
\textbf{Combined} &
\textbf{PN Embedding\textsuperscript{\textdagger}} \\
\midrule

T0
& \heatcell{51.60}{12.5}
& \heatcell{57.69}{12.5}
& \heatbest{86.98}{12.5}
& \heatcell{69.60}{12.5}
& \heatcell{86.40}{12.5}
& \heatcell{79.87}{12.5} \\

T1
& \heatcell{48.40}{12.5}
& \heatcell{60.42}{12.5}
& \heatbest{84.27}{12.5}
& \heatcell{64.47}{12.5}
& \heatcell{82.97}{12.5}
& \heatcell{74.04}{12.5} \\

T2
& \heatcell{41.90}{12.5}
& \heatcell{59.99}{12.5}
& \heatbest{80.61}{12.5}
& \heatcell{60.65}{12.5}
& \heatcell{79.20}{12.5}
& \heatcell{70.74}{12.5} \\

T3
& \heatcell{30.59}{12.5}
& \heatcell{55.28}{12.5}
& \heatbest{78.85}{12.5}
& \heatcell{56.20}{12.5}
& \heatcell{77.05}{12.5}
& \heatcell{66.65}{12.5} \\

\addlinespace[2pt]

P1
& \heatcell{50.31}{12.5}
& \heatcell{57.47}{12.5}
& \heatbest{83.27}{12.5}
& \heatcell{67.50}{12.5}
& \heatcell{83.24}{12.5}
& \heatcell{77.85}{12.5} \\

P2
& \heatcell{47.51}{12.5}
& \heatcell{53.98}{12.5}
& \heatcell{79.03}{12.5}
& \heatcell{61.83}{12.5}
& \heatbest{79.26}{12.5}
& \heatcell{72.37}{12.5} \\

\addlinespace[2pt]

P1$^\ast$
& \heatcell{29.28}{12.5}
& \heatcell{52.38}{12.5}
& \heatbest{73.34}{12.5}
& \heatcell{53.65}{12.5}
& \heatcell{72.49}{12.5}
& \heatcell{63.28}{12.5} \\

P2$^\ast$
& \heatcell{27.22}{12.5}
& \heatcell{46.22}{12.5}
& \heatbest{67.87}{12.5}
& \heatcell{46.57}{12.5}
& \heatcell{66.53}{12.5}
& \heatcell{56.86}{12.5} \\

\bottomrule
\end{tabular}

\caption{Within-language 8-way cross-family attribution accuracy (\%) on Persian outputs using logistic regression. Chance accuracy is 12.5\%.}
\label{tab:cross_family_persian}
\end{table}

\begin{table}[t]
\centering
\small
\setlength{\tabcolsep}{5pt}
\renewcommand{\arraystretch}{1.15}

\begin{tabular}{lccccc@{\hspace{7pt}}c}
\toprule
\textbf{View} &
\textbf{Behavioral} &
\textbf{Stylometric} &
\textbf{Lexical} &
\textbf{Semantic} &
\textbf{Combined} &
\textbf{PN Embedding\textsuperscript{\textdagger}} \\
\midrule

T0
& \heatcell{54.57}{12.5}
& \heatcell{59.01}{12.5}
& \heatbest{83.79}{12.5}
& \heatcell{68.48}{12.5}
& \heatcell{83.62}{12.5}
& \heatcell{83.76}{12.5} \\

T1
& \heatcell{51.92}{12.5}
& \heatcell{57.90}{12.5}
& \heatbest{80.32}{12.5}
& \heatcell{63.80}{12.5}
& \heatcell{79.93}{12.5}
& \heatcell{77.32}{12.5} \\

T2
& \heatcell{43.96}{12.5}
& \heatcell{56.82}{12.5}
& \heatbest{77.61}{12.5}
& \heatcell{59.82}{12.5}
& \heatcell{76.41}{12.5}
& \heatcell{75.42}{12.5} \\

T3
& \heatcell{29.08}{12.5}
& \heatcell{50.78}{12.5}
& \heatbest{73.98}{12.5}
& \heatcell{55.55}{12.5}
& \heatcell{72.50}{12.5}
& \heatcell{71.28}{12.5} \\

\addlinespace[2pt]

P1
& \heatcell{53.52}{12.5}
& \heatcell{57.45}{12.5}
& \heatcell{80.74}{12.5}
& \heatcell{66.96}{12.5}
& \heatbest{81.12}{12.5}
& \heatcell{81.26}{12.5} \\

P2
& \heatcell{51.99}{12.5}
& \heatcell{56.81}{12.5}
& \heatcell{78.80}{12.5}
& \heatcell{64.60}{12.5}
& \heatbest{79.46}{12.5}
& \heatcell{79.13}{12.5} \\

\addlinespace[2pt]

P1$^\ast$
& \heatcell{28.60}{12.5}
& \heatcell{49.09}{12.5}
& \heatcell{69.07}{12.5}
& \heatcell{52.39}{12.5}
& \heatbest{69.16}{12.5}
& \heatcell{68.85}{12.5} \\

P2$^\ast$
& \heatcell{27.86}{12.5}
& \heatcell{48.25}{12.5}
& \heatcell{65.85}{12.5}
& \heatcell{51.66}{12.5}
& \heatbest{67.27}{12.5}
& \heatcell{66.79}{12.5} \\

\bottomrule
\end{tabular}

\caption{Within-language 8-way cross-family attribution accuracy (\%) on Chinese outputs using logistic regression. Chance accuracy is 12.5\%.}
\label{tab:cross_family_chinese}
\end{table}

\section{Cross-Lingual Fingerprint Transfer}
\label{app:cross_lingual_transfer}

To examine whether model-specific fingerprints persist across languages, we train the attribution classifier on outputs in one language and evaluate it on outputs from another language. We consider all six directed transfers among English (EN), Persian (FA), and Mandarin Chinese (ZH). Evaluation uses the common set of 798 prompts and five prompt-disjoint folds, such that the translated counterpart of a target-language test prompt is never present in the source-language training split. Logistic-regression hyperparameters are selected using only the source-language training data. TF--IDF is excluded because its vocabulary is intrinsically tied to language and script. Table~\ref{tab:cross_lingual_average} averages accuracy over all six transfer directions.

For cross-lingual experiments, we retain the same feature definition across languages to avoid changing the classifier representation between source and target languages. Consequently, some Stylometric components, including the English-trained POS representation, are not linguistically calibrated for Persian or Chinese. We therefore treat cross-lingual Stylometric accuracy as a fixed-feature diagnostic and do not interpret it as evidence of language-general POS structure.

\begin{table}[t]
\centering
\small
\setlength{\tabcolsep}{5pt}
\renewcommand{\arraystretch}{1.15}

\begin{adjustbox}{max width=\linewidth}
\begin{tabular}{lccccc@{\hspace{7pt}}c}
\toprule
\textbf{View} &
\textbf{Behavioral} &
\textbf{Stylometric} &
\textbf{Semantic} &
\textbf{Combined} &
\textbf{Best Deployable} &
\textbf{PN Embedding\textsuperscript{\textdagger}} \\
\midrule

T0
& \heatcell{29.65}{12.5}
& \heatcell{19.74}{12.5}
& \heatcell{30.58}{12.5}
& \heatcell{33.36}{12.5}
& \heatbest{34.56}{12.5}
& \heatcell{53.03}{12.5} \\

T1
& \heatcell{27.34}{12.5}
& \heatcell{19.88}{12.5}
& \heatcell{29.80}{12.5}
& \heatcell{31.94}{12.5}
& \heatbest{33.23}{12.5}
& \heatcell{48.57}{12.5} \\

T2
& \heatcell{21.69}{12.5}
& \heatcell{20.51}{12.5}
& \heatcell{25.86}{12.5}
& \heatcell{24.99}{12.5}
& \heatbest{28.49}{12.5}
& \heatcell{44.20}{12.5} \\

T3
& \heatcell{19.42}{12.5}
& \heatcell{18.56}{12.5}
& \heatcell{24.21}{12.5}
& \heatcell{24.10}{12.5}
& \heatbest{26.66}{12.5}
& \heatcell{39.00}{12.5} \\

\addlinespace[2pt]

P1
& \heatcell{28.78}{12.5}
& \heatcell{20.58}{12.5}
& \heatcell{31.11}{12.5}
& \heatcell{33.62}{12.5}
& \heatbest{34.22}{12.5}
& \heatcell{53.34}{12.5} \\

P2
& \heatcell{29.95}{12.5}
& \heatcell{19.36}{12.5}
& \heatcell{30.60}{12.5}
& \heatcell{33.88}{12.5}
& \heatbest{35.31}{12.5}
& \heatcell{50.21}{12.5} \\

\addlinespace[2pt]

P1$^\ast$
& \heatcell{18.59}{12.5}
& \heatcell{18.65}{12.5}
& \heatcell{23.30}{12.5}
& \heatcell{23.40}{12.5}
& \heatbest{25.54}{12.5}
& \heatcell{37.85}{12.5} \\

P2$^\ast$
& \heatcell{17.54}{12.5}
& \heatcell{18.09}{12.5}
& \heatcell{23.26}{12.5}
& \heatcell{23.08}{12.5}
& \heatbest{24.48}{12.5}
& \heatcell{36.64}{12.5} \\

\bottomrule
\end{tabular}
\end{adjustbox}

\caption{Average cross-lingual 8-way attribution accuracy (\%) across the six directed language-transfer settings using logistic regression. Chance accuracy is 12.5\%.
}
\label{tab:cross_lingual_average}
\end{table}

\end{document}